%% file: main.tex
\documentclass[10pt,twocolumn,letterpaper]{article}

\usepackage[pagenumbers]{iccv} 

\input{preamble}

\definecolor{iccvblue}{rgb}{0.21,0.49,0.74}
\usepackage[pagebackref,breaklinks,colorlinks,allcolors=iccvblue]{hyperref}
\usepackage{url}
\usepackage{booktabs}
\usepackage{graphicx}
\usepackage{multirow}
\usepackage{subcaption}
\usepackage{wrapfig}
\usepackage{adjustbox}
\usepackage{enumitem}
\usepackage{algorithm}
\usepackage{algorithmic}
\usepackage{setspace}
\usepackage{bm}
\usepackage{makecell}
\usepackage{caption}
\usepackage{subcaption}

\usepackage{comment}

\usepackage{multicol}
\usepackage{threeparttable}
\usepackage{amsfonts}       
\usepackage{amssymb}     
\usepackage{amsmath}     
\usepackage{nicefrac}       
\usepackage{microtype}      
\usepackage{xcolor}         
\usepackage{bm}     
\usepackage{xfrac} 
\usepackage{array}

\newcommand{\PreserveBackslash}[1]{\let\temp=\\#1\let\\=\temp}
\newcolumntype{C}[1]{>{\PreserveBackslash\centering}p{#1}}
\newcolumntype{R}[1]{>{\PreserveBackslash\raggedleft}p{#1}}
\newcolumntype{L}[1]{>{\PreserveBackslash\raggedright}p{#1}}

\definecolor{Gray}{rgb}{0.95, 0.95, 0.95}

\def\paperID{6218} 
\def\confName{ICCV}
\def\confYear{2025}

\title{Consistency Trajectory Matching for One-Step Generative Super-Resolution}

\author{\hspace{-0.5cm}
Weiyi You\quad Mingyang Zhang\quad
Leheng Zhang\quad Xingyu Zhou\quad Kexuan Shi\quad Shuhang Gu\thanks{Corresponding author.}\\
\hspace{-0.5cm}
University of Electronic Science and Technology of China \hspace{0pt}\\
\hspace{-0.5cm}
{\tt\small \{weiyiyou.ywy, shuhanggu\}@gmail.com}\\
\small \url{https://github.com/LabShuHangGU/CTMSR}}

\begin{document}
\maketitle
\input{sec/0_abstract}

\input{sec/1_intro}

\input{sec/2_related}

\input{sec/3_methodology}

\input{sec/4_experiment}
\input{sec/5_conclusion}

{
    \small
    \bibliographystyle{ieeenat_fullname}
    \bibliography{main}
}

\input{sec/X_suppl}

\end{document}

%% file: preamble.tex
\usepackage[utf8]{inputenc}
\usepackage{amsmath, amssymb}

\usepackage[ruled, lined, linesnumbered, commentsnumbered, longend]{algorithm2e}

\usepackage{listings}
\usepackage{graphicx}
\usepackage{booktabs}
\usepackage{array}
\usepackage{amsmath}
\usepackage{amsfonts} 
\usepackage{graphicx}
\usepackage{caption}
\usepackage[export]{adjustbox}
\usepackage{makecell}
\usepackage{multirow}
\usepackage{rotating}
\usepackage{environ}
\usepackage{enumerate}

%% file: sec/0_abstract.tex
\begin{abstract}
Current diffusion-based super-resolution (SR) approaches achieve commendable performance at the cost of high inference overhead. Therefore, distillation techniques are utilized to accelerate the multi-step teacher model into one-step student model. Nevertheless, these methods significantly raise training costs and constrain the performance of the student model by the teacher model. To overcome these tough challenges, we propose \textbf{C}onsistency \textbf{T}rajectory \textbf{M}atching for \textbf{S}uper-\textbf{R}esolution (\textbf{CTMSR}), a distillation-free strategy that is able to generate photo-realistic SR results in one step. Concretely, we first formulate a Probability Flow Ordinary Differential Equation (PF-ODE) trajectory to establish a deterministic mapping from low-resolution (LR) images with noise to high-resolution (HR) images. Then we apply the Consistency Training (CT) strategy to directly learn the mapping in one step, eliminating the necessity of pre-trained diffusion model.  To further enhance the performance and better leverage the ground-truth during the training process, we aim to align the distribution of SR results more closely with that of the natural images. To this end, we propose to minimize the discrepancy between their respective PF-ODE trajectories from the LR image distribution by our meticulously designed Distribution Trajectory Matching (DTM) loss, resulting in improved realism of our recovered HR images. Comprehensive experimental results demonstrate that the proposed methods can attain comparable or even superior capabilities on both synthetic and real datasets while maintaining minimal inference latency.
\end{abstract}

%% file: sec/1_intro.tex
\section{Introduction}
\begin{figure}[t]
    \centering
    \begin{subfigure}[b]{0.45\textwidth}
        \centering
        \includegraphics[width=\textwidth]{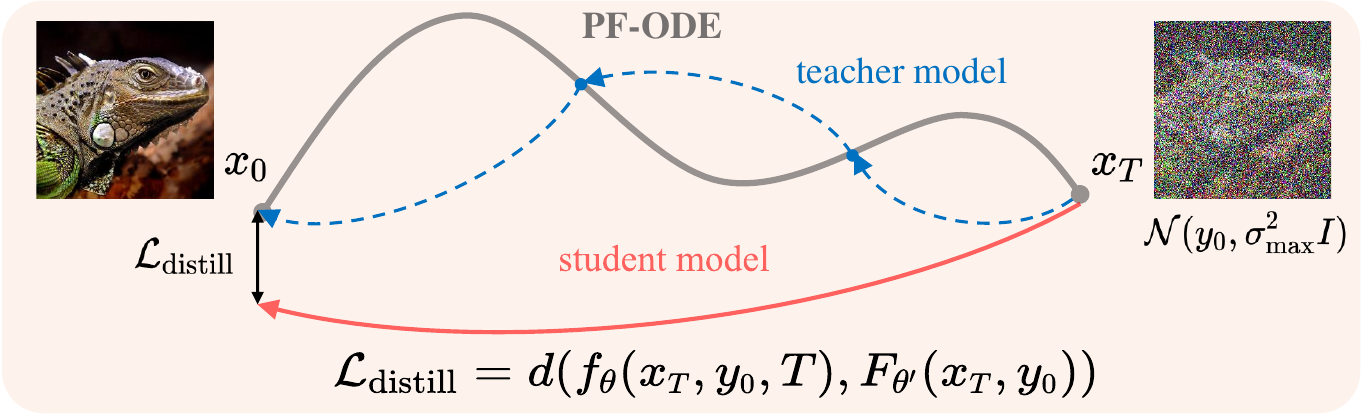}
        \caption{\textbf{Vanilla distillation.} The student model ($f_\theta$) directly learns the PF-ODE from ${x}_T$ to ${x}_0$ formed by multi-step teacher model ($F_{\theta'}$). }
        \label{pic: distillation}
    \end{subfigure}
    \hfill 
    \begin{subfigure}[b]{0.45\textwidth}
        \centering
        \includegraphics[width=\textwidth]{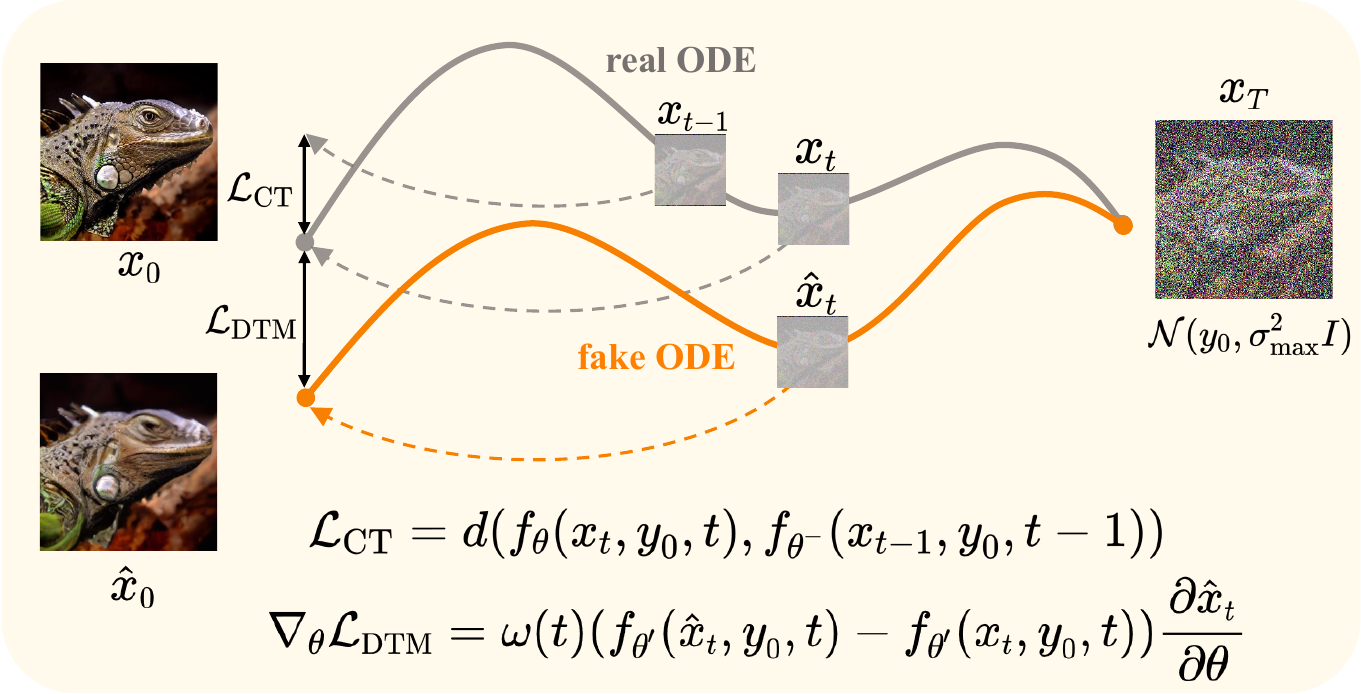}
        \caption{\textbf{Consistency Trajectory Matching for SR.} We first utilize Consistency Training to map any point on the PF-ODE to the final point $x_0$ by minimizing the distance between model outputs of two adjacent points on the PF-ODE. Based on the learned ODE, we propose DTM to match the trajectory of fake ODE with the trajectory of real ODE, making the SR results better aligned with the distribution of natural images.}
        \label{pic: CTMSR}
    \end{subfigure}

    \caption{An illustrative comparison of vanilla distillation and our proposed Consistency Trajectory Matching for SR. In contrast to vanilla distillation, Consistency Training directly learns the deterministic mapping from noisy LR distribution to the natural image distribution to achieve one-step inference and DTM is proposed to further enhance the realism of SR results.}
    \label{pic: compare}
\end{figure}
\label{sec:intro}
Single-image super-resolution (SISR) is a task of generating a high-resolution (HR) image that is in accordance with the input low-resolution (LR) image. As is known, SISR is a typical ill-posed problem in the field of low-level vision, since every LR image could consist with a number of potential HR counterparts. 
Early classical SR methods~\cite{zhang2024transcending, liang2021swinir, tian2024image, chen2023activating} restore the HR images via optimizing the Root Mean Square Error (RMSE) loss function in a supervised manner. This methodology forces the model to learn an expectation of all possible HR counterparts, which leads to blurry SR results~\cite{ledig2017photo}. While generative SR methods aim to generate the HR estimation that conforms to the natural image distribution, thus producing more photo-realistic HR images.
Recently, diffusion models~\cite{song2020denoising, ho2020denoising} have demonstrated strong capabilities in modeling complex distributions, e.g., the distribution of natural images, holding great potential for generative SR. 
Early diffusion-based SR works~\cite{saharia2022image, li2022srdiff, wang2024exploiting, kawar2022denoising} either condition the diffusion model on LR and train it as a common diffusion model (e.g., DDPM~\cite{ho2020denoising}), or leverage a pre-trained diffusion model as a prior and adjust the reverse process guided by LR images.
Though these methods yield decent results, both of them require hundreds of inference steps. 
Therefore, numerous attempts have been made to accelerate the inference speed of diffusion-based SR models.
While some studies~\cite{song2020denoising, lu2022dpm} investigated advanced inference strategies for reducing the sampling steps; \cite{yue2024resshift, luo2023image} propose to model the initial state of diffusion process as low-quality image perturbed by a slight amount of noise rather than pure noise, greatly reducing the inference steps for generative SR.
Furthermore, SinSR~\cite{wang2024sinsr} reformulates the inference process of ResShift~\cite{yue2024resshift} as Ordinary Differential Equation (ODE) and directly distills it into one step.
However, as mentioned in \cite{liu2022flow}, the performance of one-step student model is limited by the teacher model; if the ODE is not rectified to be straight during the training process of the teacher model, direct distillation could only produce sub-optimal results.
Besides, distilling the teacher model involves multi-step sampling to generate training data pairs, which greatly increases the training overhead. 
Beyond these approaches, some Stable Diffusion-based methods~\cite{xie2024addsr, wu2025one} leverage the powerful generative capabilities of pretrained Stable Diffusion (SD) and achieve impressive results in a single inference step. However, their reliance on a fixed backbone limits scalability to smaller models, restricting practical applicability.
Therefore, how to obtain a distillation-free and backbone-independent one-step generative SR model that can produce photo-realistic SR results with limited inference footprint remains a challenging problem in the literature.

In order to tackle the aforementioned issues, we propose Consistency Trajectory Matching Super-Resolution (CTMSR), an efficient generative SR approach that could produce high-perceptual-quality HR images in merely one step. 
Instead of distilling one-step model from a pre-trained generative SR model, we leverage recent advances in Consistency Training (CT) ~\cite{song2023consistency, song2023improved} and  directly learn a mapping function between LR images with noise to HR images.
The proposed CT strategy enables us to directly learn a Probability Flow Ordinary Differential Equation (PF-ODE) trajectory, therefore eliminating the limitation of pre-trained multi-step diffusion model.
Moreover, based on the learned PF-ODE trajectory, which is capable of transitioning noisy LR distribution to the natural image distribution, we propose the Distribution Trajectory Matching (DTM) loss to further improve our SR results.
The proposed DTM loss penalizes the distribution discrepancy between our SR results and high quality images in a trajectory level by matching their respective PF-ODEs from the noisy LR distribution, resulting in improved perceptual quality of our SR results.
Extensive experimental results on synthetic and real-world datasets clearly demonstrate the superiority of our methods. With less inference footprint, our proposed CTMSR is able to generate state-of-the-art photo-realistic SR results.

Our main contributions are summarized as follows:

\begin{itemize}

    \item We propose the Consistency Training for SR to directly establish a PF-ODE from the noisy LR distribution to HR distribution. This enables us to produce photo-realistic SR results in one step without the need for distillation, achieving efficiency in both training and inference.

    \item Built upon the learned PF-ODE trajectory, we propose Distribution Trajectory Matching to better align the distribution of SR results with the distribution of natural images via trajectory matching, greatly enhancing realism.

    \item We provide comprehensive experimental results on both synthetic and real-world datasets. Compared with existing methods, our CTMSR achieves comparable or even better performance while maintaining less inference latency. 
\end{itemize}

%% file: sec/2_related.tex
\section{Related Work}
\label{sec: related}

\subsection{Image Super-Resolution}
\label{subsec: sr}
Image super-resolution is a classical ill-posed problem that presents significant challenges in the field of low-level vision. Conventional SR methods~\cite{dong2012nonlocally, gu2015convolutional} recover the details of HR images via manually designing image priors guided by subjective knowledge. With the emergence of Deep Learning (DL), DL-based methods gradually dominate the realm of SR. Specifically, existing DL-based SR methods can be roughly categorized into two types: fidelity-oriented SR and generative SR. There exists numerous researches in fidelity-oriented SR~\cite{zhou2020guided, dong2015image, chen2023activating, li2023efficient, liang2021swinir, zhang2024transcending} that relies on minimization of the pixel distance (e.g., $\ell_2$ distance) between the reconstructed HR image and the ground-truth image in a supervised manner. Each of them makes efforts to improve the fidelity performance of SR from different aspects, varying from network architectures to loss functions and training strategies, and so on. Despite their success in achieving high Peak Signal-to-Noise Ratio (PSNR) scores, they inevitably produce over-smooth SR results. To overcome this challenge, generative SR methods~\cite{ledig2017photo, wang2018esrgan, wang2021real} leverage the characteristic of generative models to model the distribution of natural images, aiming to optimize the SR model at the distribution level. Among them, diffusion-based techniques demonstrate exceptional performance in enhancing perceptual quality of SR results. Early diffusion-based SR methods~\cite{saharia2022image, li2022srdiff, rombach2022high} condition the diffusion model on LR images and train it the same way as a conventional diffusion model. Alternatively, ~\cite{wang2024exploiting, kawar2022denoising, choi2021ilvr} utilize a pre-trained diffusion model as a prior and modify the reverse process based on LR images. Although these approaches yield satisfactory results, they generally require dozens or even hundreds of inference steps to generate HR images, since both methods start from an initial state of pure noise. To further enhance efficiency and tailor diffusion models more effectively for SR, ResShift~\cite{yue2024resshift} proposes modeling the initial state of the diffusion process as a LR image with a slight amount of noise rather than pure noise, thereby substantially reducing the required inference steps to 15. Additionally, SinSR~\cite{wang2024sinsr} directly distills ResShift into single step. Although the distillation method has achieved substantial reductions in inference computational expense, limitations persist. It inevitably leads to considerable training costs and restricts the performance of the student model by the limitations of the teacher model.

\subsection{Acceleration of Diffusion Models}
\label{subsec: acceleration}
Despite the strong generation capabilities manifested by diffusion models, considerable inference time overhead significantly hinders their practical application. Therefore, a range of acceleration techniques have been proposed to alleviate this issue. Certain approaches accomplish this by refining the inference process~\cite{song2020denoising, lu2022dpm, zhao2024unipc}, while several methods~\cite{karras2022elucidating, nichol2021improved} concentrate on improving the diffusion schedule. Though these methods effectively reduce the inference steps to dozens, performance deteriorates markedly when the step count falls below ten. To overcome this limitation, distillation methods~\cite{yin2024one, liu2022flow, salimans2022progressive} are proposed to further compress the steps below ten while preserving promising performance. Among them, Progressive Distillation~\cite{salimans2022progressive} effectively reduces the inference steps of student models through a multistage distillation. Nevertheless, the compounding errors at each distillation stage significantly undermine the overall performance of the student model. DMD~\cite{yin2024one} seeks to minimize the Kullback–Leibler (KL) divergence between the distribution of generated images and that of natural images by distilling the scores in pre-trained diffusion models, ultimately reducing the inference process to a single step. For SR task, the distillation approach has also been leveraged by SinSR~\cite{wang2024sinsr} to distill ResShift~\cite{yue2024resshift} into one step. In addition, Consistency Model~\cite{song2023consistency} is able to achieve promising results in 2${\sim}$4 steps, which is trained either by distillation or from scratch. Drawing inspiration from Consistency Model, we propose a distillation-free diffusion-based SR method with one-step inference in this paper.

%% file: sec/3_methodology.tex
\section{Methodology}

\subsection{Preliminaries}
\label{subsec: prelim}

\noindent\textbf{Diffusion Models.}
Diffusion models are a type of generation model that transforms the distribution of natural images (i.e., $p_\text{data}(\bm{x})$) into a Gaussian noise distribution (i.e., $\mathcal{N}(0, \sigma_\text{max}^2\bm{I})$ through a forward process and constructs a reverse sampling process from pure noise to natural images. Specifically, the forward marginal distribution is defined as:       $q(\bm{x}_t|\bm{x}_{0})=\mathcal{N}(\bm{x}_t;\bm{x}_{0},\sigma(t)^2\bm{I})$, where $\sigma(t)$ is a predefined function that controls the schedule of noise and obeys $\sigma(0) = 0$ and $\sigma(T) = \sigma_\text{max}$. To simplify the representation of $\bm{x}_t$, the forward marginal distribution can be reparameterized as:
\begin{equation}
\label{eq: forward distribution repar}
    \begin{split}
       \bm{x}_{t} = \bm{x}_{0} \ + \ \sigma(t) \bm{\epsilon},
       \quad\mathrm{where} \ \bm{\epsilon} \sim\mathcal{N}(\bm{0},\bm{I}).
    \end{split}
\end{equation}
According to \cite{karras2022elucidating, song2020score}, the forward process could be represented in the form of Stochastic Differential Equation (SDE):
\begin{equation}
\label{eq: forward sde}
\mathrm{d}\bm{x}= \dot{\sigma}(t) \ \mathrm{d}\omega_{t},
\end{equation}
where the dot denotes a time derivative and $\omega_{t}$ is the standard Wiener process. Correspondingly, an ordinary differential equation (ODE) can be employed to represent the reverse solution of this forward SDE, called the Probability Flow ODE (PF-ODE)~\cite{song2020score, lu2022dpm}:
\begin{equation}
\label{eq: reverse ode}
\mathrm{d}\bm{x}=\dot{\sigma}(t)  \ \bm{\epsilon}_{\bm{\theta}} (\bm{x}_{t}, \bm{y}_0, t)\ \mathrm{d}t ,
\end{equation}
where $\bm{\epsilon}_{\bm{\theta}}(\bm{x}_{t}, \bm{y}_0, t)$ is reparameterized by a neural network with parameter $\bm{\theta}$, aiming at predicting $\bm{\epsilon}$.

\noindent\textbf{Consistency Training.} 
With the PF-ODE formulated as Eq.~\ref{eq: reverse ode}, the Consistency Model~\cite{song2023consistency} (CM) directly estimates the solution of the PF-ODE, thus allowing for one-step generation:  

\begin{equation}
    \bm{f}_{\bm{\theta}}(\bm{x}_T,T)\approx\bm{x}_0=\bm{x}_T+\int_T^0\frac{\mathrm{d}\bm{x}_s}{\mathrm{d}s}\mathrm{d}s.
\end{equation}
Specifically, Consistency Training (CT) is proposed to train a CM that eliminates the need of pre-trained diffusion model. It first samples two adjacent points along the ODE trajectory and then minimizes the difference between model outputs corresponding to these two points. Then the training objective
\begin{equation}
\mathcal{L}(\bm{\theta},\bm{\theta}^-)=\mathbb{E}_{\bm{x},t}[d\left(f_{\bm{\theta}}(\bm{x}_{t},t),\bm{f}_{\bm{\theta}^-}(\bm{x}_{t-1},t-1)\right)]
\end{equation}
is adopted to optimize the online model $\bm{\theta}$ to approximate the target model, where $d(\cdot,\cdot)$ denotes a predefined metric function for measuring the distance between two samples and $\bm{\theta}^-$ is obtained by exponential moving average (EMA) of the parameter $\bm{\theta}$, i.e., $\bm{\theta}^- \leftarrow \mu \bm{\theta}^- \ + \ (1 - \mu)\bm{\theta}$.

\begin{figure*}[t]
    \centering
    \includegraphics[width=0.95\textwidth,keepaspectratio]{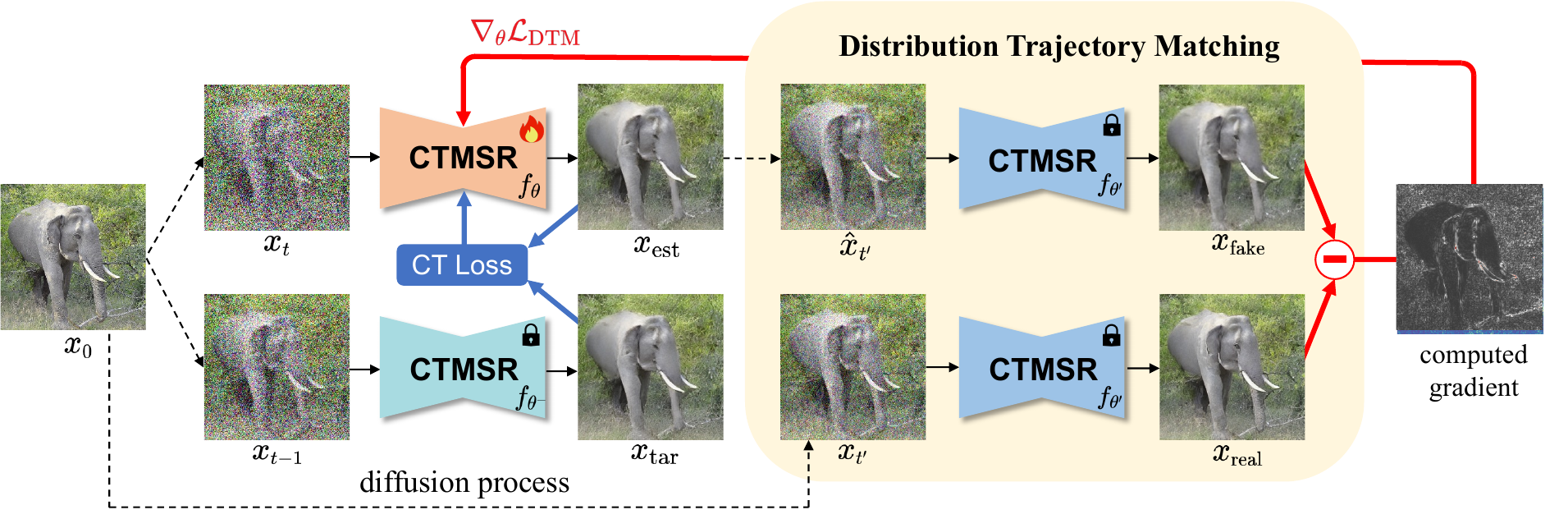}
    \captionsetup{width=1\textwidth}
    \caption{The pipeline of the proposed CTMSR. We first employ CT loss to train our CTMSR until convergence to get a pre-trained CTMSR ($f_{\theta'}$) with parameters frozen. As our pre-trained CTMSR is able to construct the PF-ODE trajectory from one distribution to another, we feed $\hat{x}_{t'}$ and $x_{t'}$ into the pre-trained CTMSR to get the trajectories of fake ODE and real ODE respectively, namely $x_\text{fake}$ and $x_\text{real}$. Then we calculate the $\nabla_\theta\mathcal{L}_{\mathrm{DTM}}$ that matches the trajectories to penalize the distribution discrepancy between our SR results and the real images in a trajectory level. With the calculated $\nabla_\theta\mathcal{L}_{\mathrm{DTM}}$ backpropagated to our training CTMSR, the realism of SR results produced by our model will be further enhanced. }
    \label{pipeline}
    \vspace{-1.5em}
\end{figure*}

\noindent\textbf{Score Distillation.} 
Score distillation methods~\cite{poole2022dreamfusion, wang2024prolificdreamer} are proposed for training a 3D generator with pre-trained image diffusion models.
Specifically, by perturbing the rendered image $\hat{z}$ with noise $\bm{\epsilon}$, the seminal work of score distillation sampling (SDS)~\cite{poole2022dreamfusion} is able to penalize the discrepancy between rendered images and the distribution captured by pre-trained diffusion model:
\begin{equation}
\label{eq: sds}
\nabla_\theta\mathcal{L}_{\mathrm{SDS}}(\hat{\bm{z}}, t, \bm{\epsilon}) = 
    \left(\bm{\epsilon}_{\bm{\phi}}(\hat{\bm{z}}_t, t) - \bm{\epsilon}\right)\frac{\partial\hat{\bm{z}}_t}{\partial\bm{\theta}},
\end{equation}
where $\hat{\bm{z}}_t$ refers to the noised version of $\hat{\bm{z}}$, $\bm{\epsilon}_{\bm{\phi}}(\cdot)$ is the pretrained diffusion model and $\bm{\theta}$ denotes the  generator parameter
.
More details about score distillation methods can be found in~\cite{poole2022dreamfusion, wang2024prolificdreamer, hertz2023delta}.
In this paper, the above idea of score distillation inspires us to align the distribution of generated images, i.e. SR outputs, with natural images through trajectory matching.
Since our CT could construct a PF-ODE trajectory between noisy LR images and high quality images, we are able to optimize the distribution discrepancy between our SR results and high quality images by matching their respective PF-ODEs from the noisy LR distribution. 

\subsection{Consistency Trajectory Matching for SR}
\label{subsec:DTM}

Current diffusion-based SR models typically rely on multi-step inference, which incurs significant time overhead. Although distillation techniques have been employed to reduce the inference steps to a single step, they still suffer from high training costs and the performance limitations imposed by the teacher model. To address these issues, we first introduce the application of CT strategy into SR model to achieve one-step inference in a distillation-free manner in Sec.~\ref{subsubsec: ctsr}. Besides, to better align the SR results with the natural image distribution, we propose Distribution Trajectory Matching in Sec.~\ref{subsubsec:DTM} to match their respective PF-ODE trajectories from the LR image distribution.

\subsubsection{Consistency Training for SR}
\label{subsubsec: ctsr}
To better leverage the prior information from LR images, we formulate the forward process tailored for SR task~\cite{yue2024resshift} based on Eq.~\ref{eq: forward distribution repar} :
\begin{equation}
\label{eq: forward distribution repar ctsr}
    \begin{split}
       \bm{x}_{t} = \bm{x}_{0} \ + \ \alpha(t) \bm{e}_{0} + \sigma(t) \bm{\epsilon},
       \quad\mathrm{where} \ \bm{\epsilon} \thicksim\mathcal{N}(\bm{0},\bm{I}),
    \end{split}
\end{equation}
where $\alpha(t)$ is a predefined function that controls the schedule of residual and obeys $\alpha(0) = 0$ and $\alpha(T) = 1$. Based on Eq.~\ref{eq: reverse ode}, we formulate the PF-ODE as: 
\begin{equation}
\label{eq: reverse ode ctsr}
\mathrm{d}\bm{x}=[\dot{\alpha}(t)  \ \bm{e}_{\bm{\theta}}(\bm{x}_{t}, \bm{y}_0, t) \ + \ \dot{\sigma}(t)  \ \bm{\epsilon}_{\bm{\theta}} (\bm{x}_{t}, \bm{y}_0, t)]\ \mathrm{d}t ,
\end{equation}
where $\bm{e}_{\bm{\theta}}(\bm{x}_{t}, \bm{y}_0, t)$ is reparameterized by a neural network with parameter $\bm{\theta}$ that aims at predicting $\bm{e}_0$. As described in Eq.~\ref{eq: reverse ode ctsr}, HR images can be restored from LR images by solving the PF-ODE from $T$ to 0. Then we introduce the consistency model $\bm{f}_{\bm{\theta}}(\bm{x}_t,t) \rightarrow \bm{x}_0$ to map any point on the PF-ODE to the final solution for $t = 0$. We parameterize the $\bm{f}_{\bm{\theta}}$ as follows:
\begin{equation}
\label{eq: model_param}
\bm{f}_{\bm{\theta}}(\bm{x}_t, \bm{y}_0, t) = c_{\text{skip}}(t) \bm{x}_t \ + \ c_{\text{out}}(t) \bm{F}_{\bm{\theta}}(\bm{x}_t, \bm{y}_0, t),
\end{equation}
where $c_{\text{skip}}(t)$ and $c_{\text{out}}(t)$ are predefined to satisfy $c_{\text{skip}}(0) = 1, c_{\text{out}}(0) = 0$  and $\bm{F}_{\bm{\theta}}$ is the actual neural network parameterized by $\bm{\theta}$. We then discretize the trajectory into $T$ intervals, with boundaries $0, 1, \ldots, T$, namely $T+1$ points on the PF-ODE trajectory. During training, we randomly select two adjacent points on the trajectory (i.e., $x_{t-1} , x_{t}$) and minimize their consistency loss $\mathcal{L}_{\text{CT}}$ as:
\begin{equation}
\mathbb{E}_{\bm{x},n}[d(f_{\bm{\theta}}(\bm{x}_{t}, \bm{y}_0, t),\bm{f}_{\bm{\theta}^-}(\bm{x}_{t-1}, \bm{y}_0,t-1))],
\end{equation}
where $\bm{\theta}^- \leftarrow \text{stopgrad}(\bm{\theta})$ according to \cite{song2023improved}. Equipped with CT strategy, our method could reconstruct the HR images through the learned PF-ODE trajectory in single-step inference. To simplify the representation, we denote $f_{\bm{\theta}}(\bm{x}_{t}, \bm{y}_0, t)$ as $\bm{x}_\text{est}$ and $\bm{f}_{\bm{\theta}^-}(\bm{x}_{t-1}, \bm{y}_0,t-1)$ as $\bm{x}_\text{tar}$, since $\bm{x}_\text{est}$ is the estimation of target $\bm{x}_\text{tar}$.

\subsubsection{Distribution Trajectory Matching}
\label{subsubsec:DTM}
Although utilizing training strategy of CT could already yield promising results in one-step inference, limitations persist. We observe that information contained in ground-truth is not effectively utilize during training, as only the point closest to $\bm{x}_0$ (i.e., $\bm{x}_1$) could directly participate in the calculation of the consistency loss with $\bm{x}_0$, while other points could only leverage $\bm{x}_0$ in a mediated way by calculating the consistency loss with the neighbouring points. Moreover, since our SR model pre-trained with $\mathcal{L}_{\text{CT}}$ is capable of estimating the PF-ODE trajectory from one distribution to another, it offers a means to optimize SR model at the distribution level. Based on these observations, we propose Distribution Trajectory Matching (DTM), a trajectory-based loss function by which we could optimize our SR model to bring the SR results closer to the natural image distribution. 
%

Firstly, we estimate the PF-ODE trajectory to the distribution of natural images, namely the \textit{real ODE}:

\vspace{-2mm}
\begin{equation}
\label{eq: DTM_1}
\bm{f}_{\bm{\theta}'}(\bm{x}_t, \bm{y}_0, t) = \bm{x}_t + \int_t^0\frac{\mathrm{d}\bm{x}_s}{\mathrm{d}s}\mathrm{d}s,
\vspace{-2mm}
\end{equation}
where
\vspace{-2mm}
\begin{equation}
\label{eq: DTM_2}
\begin{split}
\frac{\mathrm{d}\bm{x}_s}{\mathrm{d}s}  
&= \dot{\alpha}(s)  \bm{e}_{\bm{\theta}'}(\bm{x}_{s}, \bm{y}_0, s) \ + \ \dot{\sigma}(s)  \bm{\epsilon}_{\bm{\theta}'} (\bm{x}_{s}, \bm{y}_0, s) \\
&= d_{\bm{\theta}'}(\bm{x}_s, \bm{y}_0, s),
\end{split}
\vspace{-2mm}
\end{equation}
$\bm{\theta}'$ denotes the parameters of pre-trained CTMSR. In contrast to the \textit{real ODE}, we regard the SR results produced by our model as the fake distribution and construct a \textit{fake ODE} as: 
\vspace{-2mm}
\begin{equation}
\label{eq: DTM_4}
\bm{f}_{\bm{\theta}'}(\hat{\bm{x}}_t, \bm{y}_0, t) = \hat{\bm{x}}_t + \int_t^0 d_{\bm{\theta}'}(\hat{\bm{x}}_s, \bm{y}_0, s)\mathrm{d}s.
\end{equation}
Here, $\hat{\bm{x}}_{t}$ shares the same forward process of ${\bm{x}}_{t}$:
\vspace{-2mm}
\begin{equation}
\label{eq: DTM_3}
\hat{\bm{x}}_{t} = \hat{\bm{x}}_{0} \ + \ \alpha(t) \hat{\bm{e}}_{0} + \sigma(t) \bm{\epsilon},
\end{equation}

where $\bm{\epsilon} \sim\mathcal{N}(\bm{0},\bm{I})$ , $\hat{\bm{x}}_0$ is the output of the SR model (i.e., $\hat{\bm{x}}_0 = \bm{f}_{\bm{\theta}}(\hat{\bm{x}}_{t'}, \bm{y}_0, t')$) and $\hat{\bm{e}}_{0} = \bm{y}_0 - \hat{\bm{x}}_0$.

To bring the fake distribution closer to the real distribution, we propose to align the trajectory from $\hat{\bm{x}}_t$ to the fake distribution with the trajectory from $\bm{x}_t$ to the real distribution as illustrated in Figure~\ref{pic: CTMSR}. To be specific, we expect to minimize the Distribution Trajectory Distance (DTD) between $\bm{f}_{\bm{\theta}'}(\bm{x}_t, \bm{y}_0, t)$ and $\bm{f}_{\bm{\theta}'}(\hat{\bm{x}}_t, \bm{y}_0, t)$, with the corresponding loss function as follows:
\begin{equation}
\label{eq: DTM_5}
\mathcal{L}_{\mathrm{DTD}} = \mathbb{E}_{\bm{x},t}\left\| \omega(t) [\bm{f}_{\bm{\theta}'}(\hat{\bm{x}}_t, \bm{y}_0, t) - \bm{f}_{\bm{\theta}'}(\bm{x}_t, \bm{y}_0, t) ]\right\|_2^2,
\end{equation}
where $\omega(t)$  is a weighting function that depends on $t$. We can further expand this equation into the following form based on Eq.~\ref{eq: DTM_1},~\ref{eq: DTM_2},~\ref{eq: DTM_4}:
\begin{equation}
\label{eq: DTM_6}
\begin{split}
\mathcal{L}_{\mathrm{DTD}} &= 
\mathbb{E}_{\bm{x},t}\| \omega(t) [(\hat{\bm{x}}_t - \bm{x}_t) + \\ &(\int_t^0 [d_{\bm{\theta}'}(\hat{\bm{x}}_s, \bm{y}_0, s) - d_{\bm{\theta}'}({\bm{x}}_s, \bm{y}_0, s)]\mathrm{d}s)]\|_2^2,
\end{split}
\end{equation}
where $t \in [T_\text{min}, T_\text{max}]$. In Eq.~\ref{eq: DTM_6}, the first term represents sampling points at time $t$ along both trajectories and minimizing the distance between them; the second term ensures that the directions of all subsequent points on the two paths, before time $t$, remain consistent, which implicitly minimizes the distance between these points. Therefore, we could match these two trajectories from time $t$ to 0 by minimizing $\mathcal{L}_{\mathrm{DTD}}$, resulting in a better alignment between the SR results and natural images at the distribution level.
Inspired by \cite{poole2022dreamfusion, wang2024prolificdreamer}, we minimize the $\mathcal{L}_{\mathrm{DTD}}$ to eventually get $\bm{\theta}^*=\arg\min_{\bm{\theta}}\mathcal{L}_{\mathrm{DTD}}$ by exclusively updating $\bm{\theta}$ while keeping $\bm{\theta}'$ fixed. And the gradient of $\mathcal{L}_{\mathrm{DTD}}$ with respect to the parameters $\bm{\theta}$, $\nabla_{\bm{\theta}}\mathcal{L}_{\mathrm{DTD}}$, is given by:
\vspace{-1mm}
\begin{equation}
\omega(t)\left(\bm{f}_{\bm{\theta}'}(\hat{\bm{x}}_t, \bm{y}_0, t) - \bm{f}_{\bm{\theta}'}(\bm{x}_t, \bm{y}_0, t)\right)\frac{\partial \bm{f}_{\bm{\theta}'}(\hat{\bm{x}}_t, \bm{y}_0, t)}{\partial \hat{\bm{x}}_t}\frac{\partial\hat{\bm{x}}_t}{\partial\theta}.
\vspace{3mm}
\end{equation}

\begin{algorithm}[t]
    \caption{Overall training procedure of CTMSR.}
    \label{alg: overall training process}
    \begin{algorithmic}[1]
    \setstretch{1.10}
        \REQUIRE training CTMSR $\bm{f}_{\bm{\theta}}(\cdot)$
        \REQUIRE Paired training dataset $(X, Y)$
        \STATE $\textbf{Stage 1: Consistency Training for One-Step SR}$
        \STATE $k \leftarrow 0$
        \WHILE{not converged}
            \STATE $\bm{\theta}^- \leftarrow \text{stopgrad}(\bm{\theta})$
            \STATE sample $\bm{x}_0, \bm{y}_0 \sim (X, Y)$
            \STATE sample $t \sim U(1, T(k))$
            \STATE compute $\bm{x}_{t-1}, \bm{x}_{t}$ using Eq.~\ref{eq: forward distribution repar ctsr}
            \STATE $\mathcal{L}_\text{CT} = d(\bm{f}_{\bm{\theta}}(\bm{x}_{t}, \bm{y}_0, t),\bm{f}_{\bm{\theta}^-}(\bm{x}_{t-1}, \bm{y}_0,t-1))$
                             
            \STATE Take a gradient descent step on $\nabla_\theta \mathcal{L}_\text{CT}$
            \STATE  $k \leftarrow k +1$
        \ENDWHILE
        
        \STATE $\textbf{Stage 2: Distribution Trajectory Matching}$
        \STATE $\bm{\theta}' \leftarrow \text{stopgrad}(\bm{\theta})$
        \WHILE{not converged}
            \STATE sample $\bm{x}_0, \bm{y}_0 \sim (X, Y)$
            \STATE sample $t' \sim U(1, T(k))$
            \STATE compute $\bm{x}_{t'}$ using Eq.~\ref{eq: forward distribution repar ctsr}
            \STATE $\hat{\bm{x}}_0 = \bm{f}_{\bm{\theta}}(\bm{x}_{t'}, \bm{y}_0, t')$
            \STATE sample $t \sim U(T_\text{min}, T_\text{max})$
            \STATE compute $\bm{x}_{t}, \hat{\bm{x}}_{t}$ using Eq.~\ref{eq: forward distribution repar ctsr}
            

            \STATE $\nabla_\theta\mathcal{L}_{\mathrm{DTM}}=\left(\bm{f}_{\bm{\theta}'}(\hat{\bm{x}}_t, \bm{y}_0, t) - \bm{f}_{\bm{\theta}'}(\bm{x}_t, \bm{y}_0, t)\right)\frac{\partial\hat{\bm{x}}_t}{\partial\theta}$
            \STATE Take a gradient descent step on $\nabla_\theta \mathcal{L}_\text{CT} + \nabla_\theta \mathcal{L}_\text{DTM}$
            \STATE  $k \leftarrow k +1$
        \ENDWHILE
        \RETURN Converged CTMSR $f_\theta(\cdot)$.
    \end{algorithmic}
\end{algorithm}
\vspace{-1.5em}

\begin{table*}[t]

\begin{threeparttable}
    
    \centering

    \begin{tabular}{@{}C{3.2cm}@{}|
@{}C{1.9cm}@{} @{}C{2.0cm}@{} @{}C{2.0cm}@{} @{}C{2.1cm}@{} @{}C{2.1cm}@{} @{}C{2.1cm}@{} @{}C{1.8cm}@{}}
        \Xhline{0.8pt}
        \multirow{2}*{Methods} & \multicolumn{7}{c}{Metrics} \\
        \Xcline{2-8}{0.4pt}
        & PSNR$\uparrow$ & SSIM$\uparrow$ & LPIPS$\downarrow$ & CLIPIQA$\uparrow$ & MUSIQ$\uparrow$ & MANIQA$\uparrow$ & NIQE$\downarrow$ \\
        \Xhline{0.4pt}
        ESRGAN~\cite{wang2018esrgan}       & 20.67 & 0.448 & 0.485 & 0.451 & 43.615 & 0.3212 & 8.33 \\
        BSRGAN~\cite{zhang2021designing}      & 24.42 & 0.659 & 0.259 & 0.581 & \underline{54.697} & 0.3865 & 6.08 \\
        SwinIR~\cite{liang2021swinir}      & 23.99 & 0.667 & 0.238 & 0.564 & 53.790 & 0.3882 & \underline{5.89} \\
        RealESRGAN~\cite{wang2021real}    & 24.04 & 0.665 & 0.254 & 0.523 & 52.538 & 0.3689 &  6.07 \\
        \Xhline{0.4pt}
        StableSR-200~\cite{wang2024exploiting}  & 22.19 & 0.574 & 0.318 & 0.580 & 49.885 & 0.3684 & 7.10 \\
        
        LDM-15~\cite{rombach2022high}      & 24.85 & 0.668 & 0.269 & 0.510 & 46.639 & 0.3305 & 7.21 \\

        ResShift-15~\cite{yue2024resshift}   & \underline{24.94} & \underline{0.674} & 0.237 & 0.586 & 53.182 & \underline{0.4191}  & 6.88 \\

        ResShift-4~\cite{yue2024resshift}    & \textbf{25.02} & \textbf{0.683} & \underline{0.208} & 0.600 & 52.019 & 0.3885  & 7.34 \\

        SinSR-1~\cite{wang2024sinsr}    & 24.70 & 0.663 & 0.218 & \underline{0.611} & 53.632 & 0.4161 & 6.29 \\

        {CTMSR-1 (ours)}    & 24.73 & 0.666 & \textbf{0.197} & \textbf{0.691} & \textbf{60.142} & \textbf{0.4859} & \textbf{5.66} \\
        \Xhline{0.8pt}

    \end{tabular}
    \caption{Quantitative results of models on \textit{ImageNet-Test}. The best and second best results are highlighted in \textbf{bold} and \underline{underline}. ("-N" behind the method name represents the number of inference steps)}
    \label{tab: imagenet_test}
    
\end{threeparttable}
\end{table*}

\begin{figure*}[t]
    \centering
    \includegraphics[width=0.95\textwidth,keepaspectratio]{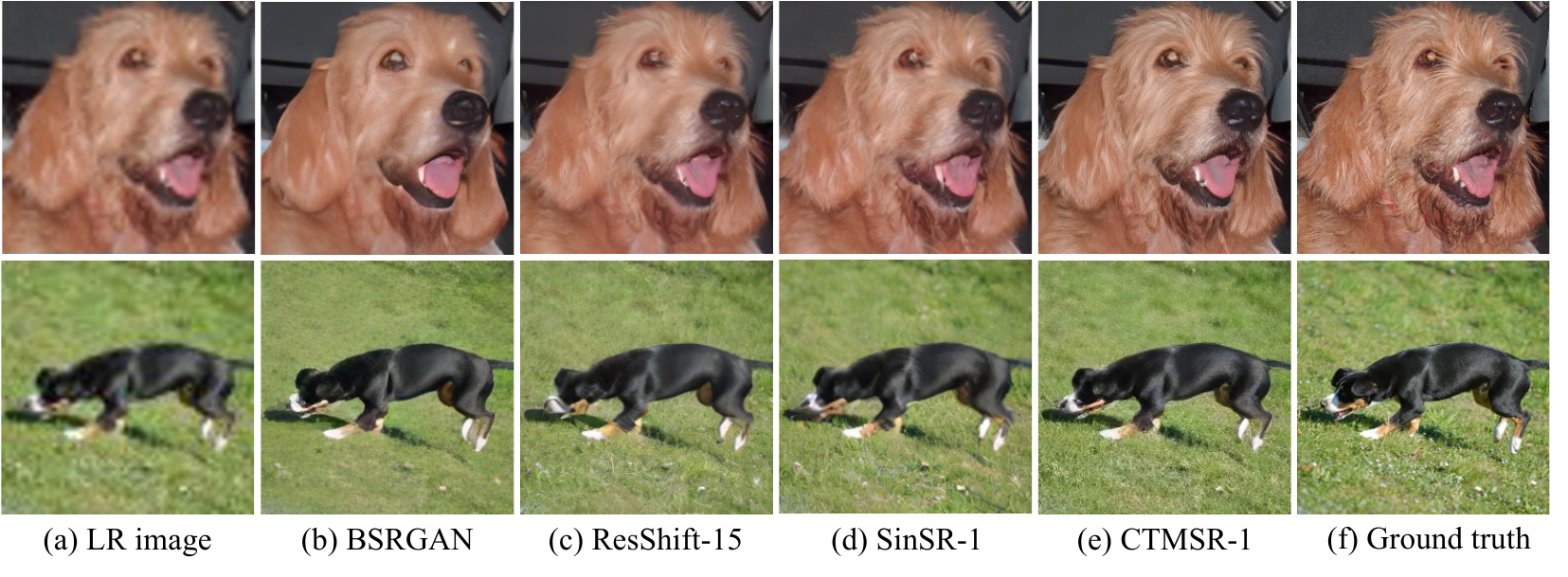}
    \captionsetup{width=0.95\textwidth}
    \vspace{-1em}
    \caption{Visual comparisons of different methods on two synthetic examples of the \textit{ImageNet-Test} dataset.}
    \label{pic: compare_imagenet}
\end{figure*}

\vspace{2mm}
In practice, calculating the U-Net Jacobian term is computationally expensive, as it involves backpropagating through the U-Net of our model. Recent studies~\cite{poole2022dreamfusion, wang2024prolificdreamer} have shown that neglecting the Jacobian term leads to more effective gradient for optimization. Inspired by this observation, we omit the differentiation through the pre-trained SR model to obtain the Distribution Trajectory Matching (DTM),
\vspace{-3mm}
\begin{equation}
    \nabla_\theta\mathcal{L}_{\mathrm{DTM}} = 
    \omega(t)\left(\bm{f}_{\bm{\theta}'}(\hat{\bm{x}}_t, \bm{y}_0, t) - \bm{f}_{\bm{\theta}'}(\bm{x}_t, \bm{y}_0, t)\right)\frac{\partial\hat{\bm{x}}_t}{\partial\theta}.
\end{equation}
In practice, we formulate $\omega(t)$ as :
\begin{equation}
\label{eq: improved weight}
    \omega(t) = \frac{CS}{\|\hat{\bm{x}}_0 - \bm{x}_0\|_1},
\end{equation}
where $S$ is the number of spatial locations and $C$ is the number of channels. The above DTM further improves the performance of our CTMSR by matching the trajectories of \textit{real ODE} and \textit{fake ODE}. We validate the effectiveness of DTM in ablation study in Sec.~\ref{subsec: ablation}. The overall of our methods is summarized in Algorithm \ref{alg: overall training process}.

\begin{table*}[t]

    \centering

    \begin{tabular}{@{}C{3.2cm}@{}|
@{}C{1.8cm}@{} @{}C{1.8cm}@{} @{}C{1.8cm}@{} @{}C{1.5cm}@{}|@{}C{1.8cm}@{} @{}C{1.8cm}@{} @{}C{1.8cm}@{} @{}C{1.5cm}@{}}
        \Xhline{0.8pt}
        \multirow{2}{*}{Methods} & \multicolumn{4}{c|}{RealSR} & \multicolumn{4}{c}{RealSet65} \\
        \Xcline{2-9}{0.4pt}
        & CLIPIQA$\uparrow$ & MUSIQ$\uparrow$ & MANIQA $\uparrow$ & NIQE$\downarrow$ & CLIPIQA$\uparrow$ & MUSIQ$\uparrow$ & MANIQA $\uparrow$ & NIQE$\downarrow$ \\
        \Xhline{0.4pt}
        StableSR-200~\cite{wang2024exploiting}         & 0.4124 & 48.346 & 0.3021 & \underline{5.87} & 0.4488 & 48.740 & 0.3097 & \underline{5.75} \\
        LDM-15~\cite{rombach2022high}         & 0.3748 & 48.698 & 0.2655 & 6.22 & 0.4313  & 48.602 & 0.2693 & 6.47  \\
        ResShift-15~\cite{yue2024resshift}     & 0.5709 & 57.769 & 0.3691 & 5.93 & 0.6309 & 59.319 & 0.3916 & 5.96\\
        ResShift-4~\cite{yue2024resshift} & 0.5646 & 55.189 & 0.3337 & 6.93 & 0.6188 & 58.516  & 0.3526 & 6.46\\
        SinSR-1~\cite{wang2024sinsr}    & \textbf{0.6627} & \underline{59.344} & \underline{0.4058} & 6.26 & \textbf{0.7164} & \underline{62.751} & \underline{0.4358} & 5.94\\
        CTMSR-1 (ours)   & \underline{0.6449} & \textbf{64.796} & \textbf{0.4157} & \textbf{4.65} & \underline{0.6893} & \textbf{67.173} & \textbf{0.4360} & \textbf{4.51} \\
        \Xhline{0.8pt}
    \end{tabular}
    \caption{Quantitative results of models on two real-world datasets. The best and second best results are highlighted in bold and underline.}
    \label{tab: real_test}
\end{table*}

\begin{figure*}[t]
    \centering
    \includegraphics[width=0.9\textwidth,keepaspectratio]{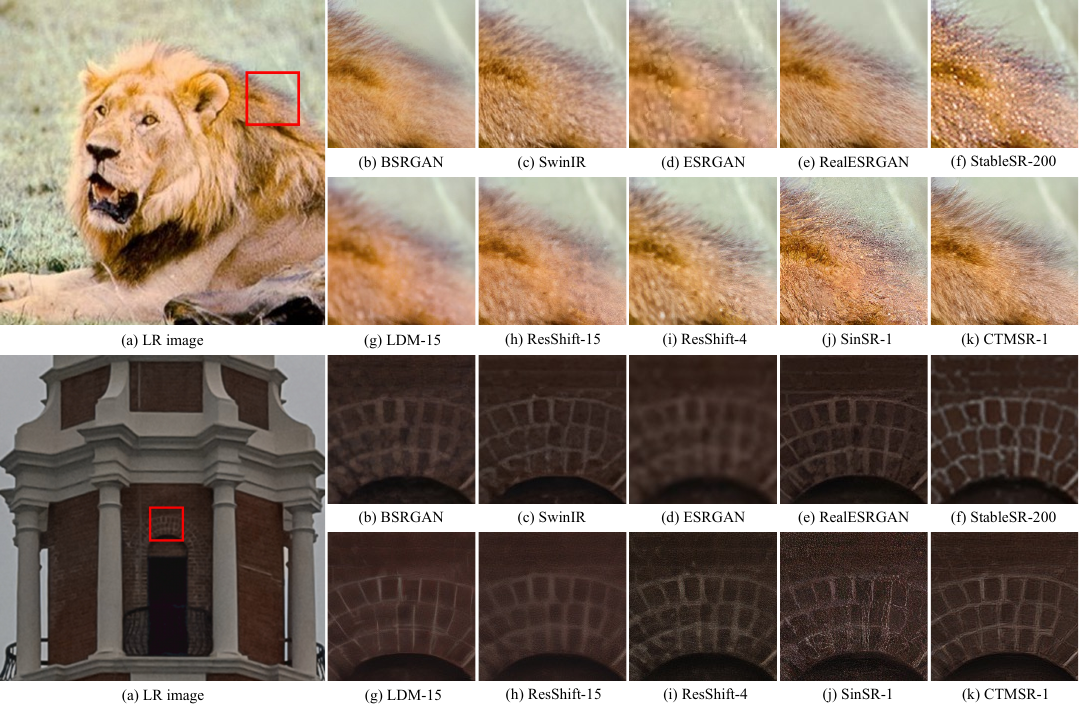}
    \captionsetup{width=0.94\textwidth}
    \vspace{-1em}
    \caption{Visual comparisons of different methods on two examples of real-world datasets. Please zoom in for more details. }
    \label{pic: visual_real}
    \vspace{-0.7em}
\end{figure*}

\subsection{Implementation details}
\textbf{Network architectures.} Analogous to ResShift~\cite{yue2024resshift}, we adopt the UNet structure with Swin Transformer~\cite{zeyde2012single} block for our CTMSR. 
While as our Consistency Training for SR and Distribution Trajectory Matching techniques could effectively capture the transition from noisy LR distribution to the natural image distribution, we do not need to rely on the encoder and decoder of pre-trained VQ-GAN model~\cite{esser2021taming} as in \cite{yue2024resshift}.
For the pursuit of efficient generative SR, we adopt tailored architecture for SR with pixel unshuffle operation and nearest neighbor upsampling, training all the parameters in the network from scratch.
More details about our network architecture can be found in the supplementary file.

\noindent\textbf{Metric function.} 
As for metric function, we adopt widely used Learned Perceptual Image Patch Similarity (LPIPS, \cite{zhang2018unreasonable}) and Charbonnier~\cite{charbonnier1997deterministic} metrics. In practice, we configure the metric function as the weighted combination of these two metrics for optimal performance:
\begin{equation}
    d(x,y) = \lambda_{1} \cdot \text{LPIPS}(x,y) + \lambda_{2} \cdot \text{Charbonnier}(x,y).
\end{equation}
In practice, we set $\lambda_{1} = 0.5$ and $\lambda_{2} = 0.5$. More implementation details are included in the supplementary materials.

%% file: sec/4_experiment.tex
\section{Experiments}
\subsection{Experimental Settings}

\noindent\textbf{Training details.} Following \cite{yue2024resshift, rombach2022high}, we randomly crop $256\times 256$ patches from the training set of ImageNet~\cite{deng2009imagenet} as our HR training data. LR images are synthesized using the degradation pipeline of RealESRGAN~\cite{wang2021real}. In the process of training, we first train our model with CT strategy for 500K iterations with fixed learning rate of 5e-5 and batch-size of 32. Then, we freeze the pre-trained model as $\bm{f}_{\bm{\theta}'}$ and further optimize $\bm{f}_{\bm{\theta}}$ with $\mathcal{L}_{\mathrm{DTM}}$ and $\mathcal{L}_{\mathrm{CT}}$ for another 2K iterations with learning rate of 5e-5.

\noindent\textbf{Testing details.} We utilize the dataset \textit{ImageNet-Test} that includes 3,000 paired images randomly selected from the validation set of ImageNet \cite{deng2009imagenet} as our main dataset following the setting in \cite{yue2024resshift}. Additionally, we adopt two real-world datasets, \textit{RealSR}~\cite{cai2019toward} and \textit{RealSet65}~\cite{yue2024resshift}, to evaluate the generalizability of our model on real-world data. To comprehensively evaluate the performance of various methods, we utilize a series of full-reference and non-reference metrics. As for full-reference metrics, PSNR and SSIM~\cite{wang2004image} are used to measure the fidelity, while LPIPS~\cite{zhang2018unreasonable}, is used to measure the perceptual quality. PSNR and SSIM are evaluated on the Y channel in the YCbCr color space. The non-reference metrics consist of NIQE~\cite{zhang2015feature}, CLIPIQA ~\cite{wang2023exploring}, MANIQA~\cite{yang2022maniqa} and MUSIQ~\cite{ke2021musiq}. NIQE assesses image quality by analyzing statistical features. MUSIQ utilizes Transformers to capture multi-scale distortions. MANIQA incorporates attention mechanisms for quality evaluation, and CLIPIQA leverages pre-trained models, such as CLIP, to align quality assessments with human perception.

\subsection{Experimental Results}
\noindent\textbf{Evaluation on testing datasets.} To demonstrate the superiority of our approach, we compare our approach with several representative SR methods, including diffusion-based methods and GAN-based methods. The diffusion-based methods incorporate StableSR~\cite{wang2024exploiting}, LDM~\cite{rombach2022high}, ResShift~\cite{yue2024resshift} and SinSR~\cite{wang2024sinsr}. Other prominent GAN-based methods encompass ESRGAN~\cite{wang2018esrgan}, BSRGAN~\cite{zhang2021designing}, SwinIR~\cite{liang2021swinir}, RealESRGAN~\cite{wang2021real}. All the test results of the compared methods are evaluated based on their released codes and pre-trained model weights.
The quantitative comparisons among various approaches are presented in Table \ref{tab: imagenet_test} and Table \ref{tab: real_test}. We can observe that our method achieves either the best or second-best performance on the perceptual quality metrics across all datasets. Specifically, on the synthetic dataset, CTMSR achieves the best performance on both reference-based and non-reference perceptual quality metrics, with only slightly lower scores on fidelity metrics PSNR and SSIM. As for real-world datasets, CTMSR achieves either the best or comparable performance across the non-reference metrics. Notably, in terms of MUSIQ, our method outperforms SinSR by 5.452 and 4.422 on the RealSR and RealSet datasets, respectively. Figure~\ref{pic: compare_imagenet} and~\ref{pic: visual_real} illustrate some visual comparisons on synthetic datasets and real-world datasets, where it can be observed that our method generates more detailed and realistic textures without noticeable artifacts.

\begin{table}[t]
\footnotesize
\setlength{\tabcolsep}{5pt}

\begin{center}
\begin{tabular}{@{}C{1.8cm}@{}|@{}C{1.4cm}@{}|@{}C{1.5cm}@{}@{}C{1.5cm}@{}@{}C{1.5cm}@{}}
\Xhline{0.8pt}
\multicolumn{1}{c|}{Methods} & \multicolumn{1}{c|}{Runtime} & LPIPS$\downarrow$ & MUSIQ$\uparrow$ & CLIPIQA$\uparrow$ \\  
\Xhline{0.4pt}
StableSR-200      &  12889 & 0.3184 & 49.885 & 0.5801 \\

LDM-15      & 223 & 0.2685 & 46.639 & 0.5095 \\
ResShift-15 & 689  & 0.2371 & 53.182 & 0.5860 \\ 
ResShift-4 & 210  & 0.2075 & 52.019 & 0.6003 \\ 
SinSR-1  & 65 & 0.2183 & 53.632 & 0.6113 \\  
CTMSR-1 & \textbf{48} & \textbf{0.1969} & \textbf{60.142} & \textbf{0.6913} \\
  
\Xhline{0.8pt}
\end{tabular}
\vspace{-0.2em}
\caption{Computational efficiency and performance comparisons with diffusion-based methods. We test the runtime (ms) on $64\times 64$ input images using single RTX3090 GPU and present several perceptual metrics evaluated on \textit{ImageNet-Test}.}
\label{tab: efficiency comparison}
\end{center}
\vspace{-2.8em}
\end{table}

\noindent\textbf{Evaluation of efficiency.} 
We measure the inference time and several perceptual quality metrics of CTMSR compared with diffusion-based approaches. Due to the reduction of inference steps to a single step, our method exhibits a significant advantage in inference latency over the multi-step approaches. As shown in Table~\ref{tab: efficiency comparison}, the inference time of our method is 22.9\% of that of ResShift-4, 6.9\% of ResShift-15, and 22.8\% of LDM-15. Despite this substantial reduction in inference time, our method still demonstrates remarkable performance superiority. Besides, compared to SinSR that also enables one-step inference, our method achieves superior performance with less inference latency, even without employing the distillation techniques. These results strongly validate that our method outperforms other diffusion-based methods in terms of both performance and efficiency.

\subsection{Ablation study}
\label{subsec: ablation}
\noindent \textbf{Effectiveness of DTM.} To enhance the alignment of SR results with the distribution of natural images, we propose DTM to perform optimization at the distribution level by matching their respective PF-ODE trajectories. In order to validate the effectiveness of DTM, we finetune the pre-trained CTMSR for another 10K iterations using $\mathcal{L}_{\mathrm{CT}}$ and $\mathcal{L}_{\mathrm{CT}}$ combined with $\mathcal{L}_{\mathrm{DTM}}$ separately. As shown in Table \ref{tab: DTM}, DTM improves CTMSR by a large margin in perceptual quality, with enhancements of 0.0821 in CLIPIQA and 3.492. Besides, it also achieves a slight improvement in fidelity. We attribute these performance improvements to the exceptional distribution matching capabilities of DTM. Based on the ablation study, we conclude that DTM effectively aligns the distribution of SR results with the distribution of natural images via trajectory matching.

\noindent\textbf{Comparison with SDS.}
To further verify that trajectory matching is more effective than score distillation~\cite{poole2022dreamfusion} for optimizing distribution discrepancy in the SR task, we also train our model with the following SDS loss:
\begin{equation}
   \nabla_\theta\mathcal{L}_{\mathrm{SDS}} =  \omega(t)\left(\bm{f}_{\bm{\theta}'}(\hat{\bm{x}}_t, \bm{y}_0, t) - \bm{x}_0 \right)\frac{\partial\hat{\bm{x}}_t}{\partial\theta}.
\end{equation}
The above equation slightly differs from the original SDS formulation~\cite{poole2022dreamfusion} because CTMSR predicts $\bm{x}_0$, whereas SDS predicts $\epsilon_t$. 
Similarly, we finetune the pre-trained CTMSR for another 5K iterations using $\mathcal{L}_{\mathrm{SDS}}$ combined with $\mathcal{L}_{\mathrm{CT}}$. As shown in Table~\ref{tab: DTM}, though SDS could also improvement non-reference perceptual quality metrics of the consistency training strategy, 
it leads to significant deterioration in terms of fidelity. In contrast, DTM achieves consistent advancements across all the metrics, delivering results that significantly outperform SDS.
Some visual examples of our ablation study can be found in Figure~\ref{pic: ablation}. More experimental results are provided in the supplementary material.

\begin{table}[t]
\footnotesize
\setlength{\tabcolsep}{5pt}



\begin{center}
\begin{tabular}{@{}C{2.5cm}@{}|@{}C{1.2cm}@{}@{}C{1.2cm}@{}@{}C{1.5cm}@{}@{}C{1.5cm}@{}}
\Xhline{0.8pt}
\multicolumn{1}{c|}{Methods} & PSNR$\uparrow$ & LPIPS$\downarrow$ & CLIPIQA$\uparrow$ & MUSIQ$\uparrow$ \\  
\Xhline{0.4pt}

CTMSR (w/o DTM) & 24.71 &  0.2004 & 0.6092 & 56.650\\
CTMSR (w/ SDS) & 23.17 & 0.2545 & 0.6292 & 58.188 \\
CTMSR (w/ DTM) & \textbf{24.73} & \textbf{0.1969} & \textbf{0.6913} & \textbf{60.142} \\

\Xhline{0.8pt}

\end{tabular}
\vspace{-0.5em}
\caption{A comparison between DTM and SDS. We evaluate their performance on \textit{ImageNet-Test}.}
\label{tab: DTM}
\end{center}
\vspace{-1.3em}
\end{table}

\begin{figure}[t]
    \centering
    \includegraphics[width=0.45\textwidth,keepaspectratio]{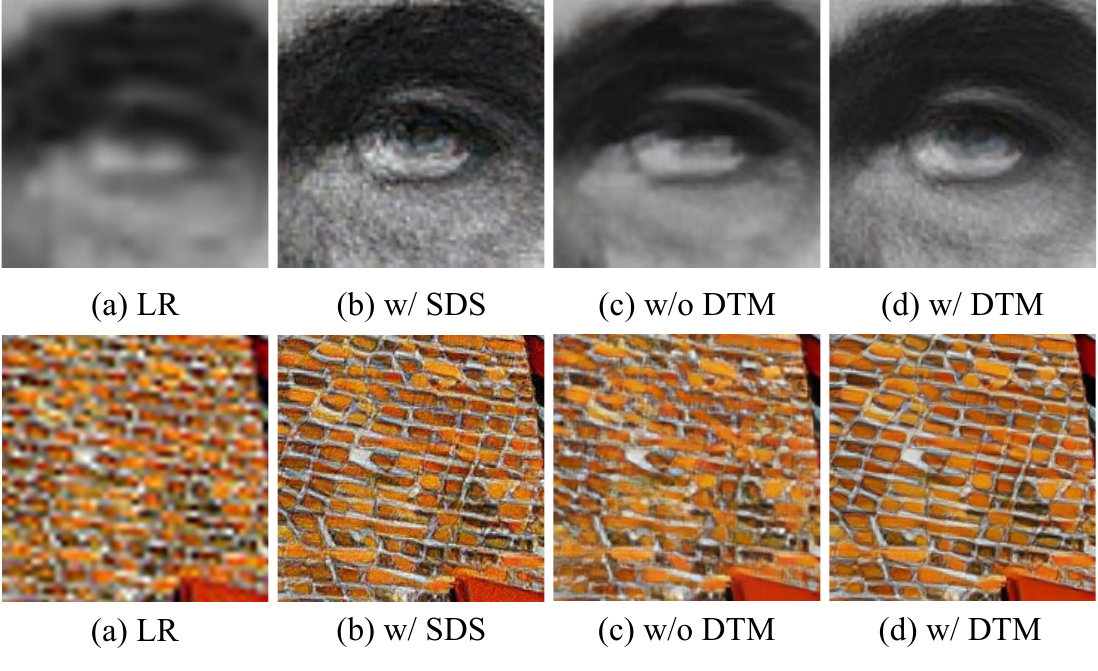}
    \captionsetup{width=0.45\textwidth}
    \vspace{-0.5em}
    \caption{A visual comparison between the impact of DTM and SDS. It can be observed that DTM restores more details and produces fewer artifacts compared to the other two methods.}
    \label{pic: ablation}
    \vspace{-1em}
\end{figure}

%% file: sec/5_conclusion.tex
\section{Conclusion}
In this paper, we propose Consistency Trajectory Matching for Super-Resolution (CTMSR), an efficient method that enables generating high-realism SR results with only one inference step without the need for distillation. We first introduce the Consistency Training for SR to directly learn the deterministic mapping between the LR images perturbed with noise to HR images, thereby establishing a PF-ODE trajectory. To better align the distribution of SR results with the distribution of natural images, we propose Distribution Trajectory Matching (DTM) that matches their respective trajectories from LR distribution based on the learned PF-ODE, resulting in significant enhancements in the realism of SR results. Extensive experimental results demonstrate that our method achieves comparable or even better performance compared to existing diffusion-based methods while maintaining the fastest inference speed.

\vspace{\baselineskip}
\noindent\textbf{Acknowledgement.} This work was supported by
 National Natural Science Foundation of China (No.62476051) and Sichuan Natural Science Foundation (No.2024NSFTD0041).

%% file: sec/X_suppl.tex
\clearpage
\setcounter{page}{1}
\setcounter{section}{0}
\maketitlesupplementary

In the supplementary materials, we introduce more details of our implementation, more experimental results and more visual comparisons.

\section*{A. Implementation Details}
\label{sec: implementation}
\subsection*{A.1. Noise and Residual Schedules}
Following \cite{karras2022elucidating}, we design the schedule for $\sigma(t)$ as follows:
\begin{equation}
\sigma(t) = \sigma_\text{max}  \cdot \left( \frac{t}{T}\right)^{\rho_{n}},
\end{equation}
where $\sigma_\text{max}$ denotes the highest noise level and $\rho_{n}$ controls the speed of noise growth; a larger $\rho_{n}$ leads to faster growth in the earlier stages and slower growth in the later stages, and vice versa. Similarly, we also design a schedule for $\alpha(t)$:
\begin{equation}
\alpha(t) = \left( \frac{t}{T}\right)^{\rho_{r}},
\end{equation}
where $\rho_{r}$ serves a role identical to that of $\rho_{n}$. In practice, we adopt the linear schedule by setting $\rho_n = 1$ and $\rho_r = 1$. 

\subsection*{A.2. Step Schedule}
We design a step schedule for Consistency Training of our SR model that adjusts the number of steps with the growth of training iterations. In contrast to ~\cite{song2023improved, song2023consistency}, we utilize a linearly decreasing curriculum for the total steps $T$, rather than an increasing one. Specifically, the curriculum is formulated as follows:
\begin{equation}
T(k) = \text{max}(s_0 -  \lfloor \frac{k}{K'} \rfloor, s_1), \quad K' = \left \lfloor \frac{K}{s_0 - s_1 + 1} \right \rfloor,
\end{equation}
where $k$ denotes the training iteration, $s_0$ denotes the initial steps, $s_1$ denotes the final steps and $K$ denotes the total iterations. We empirically discover that the decreasing step schedule could produce better results and achieve faster convergence with $s_0 = 4, s_1=3$.

\subsection*{A.3. Training Details of Distribution Trajectory Matching}
To stabilize the training of DTM, we propose to periodically update $\bm{f}_{\bm{\theta}'}$. Specifically, we update $\bm{f}_{\bm{\theta}'}$ with the parameters of $\bm{f}_{\bm{\theta}}$ every 1k iterations during the training stage of DTM.
Algorithm~\ref{alg: DTM loss} shows the details of the overall training process of CTMSR and Algorithm ~\ref{alg: DTM loss} shows the implementation of Distribution Trajectory Matching loss.


\begin{algorithm}[t]
    \caption{Overall training procedure of CTMSR.}
    \label{alg: overall training process}
    \vspace{-0.25em}
    \begin{algorithmic}[1]
    \setstretch{1.10}
        \REQUIRE training CTMSR $\bm{f}_{\bm{\theta}}(\cdot)$
        \REQUIRE Paired training dataset $(X, Y)$
        \STATE $\textbf{Stage 1: Consistency Training for One-Step SR}$
        \STATE $k \leftarrow 0$
        \WHILE{not converged}
            \STATE $\bm{\theta}^- \leftarrow \text{stopgrad}(\bm{\theta})$
            \STATE sample $\bm{x}_0, \bm{y}_0 \sim (X, Y)$
            \STATE sample $t \sim U(0, T(k)-1)$
            \STATE compute $\bm{x}_t, \bm{x}_{t-1}$ using Eq.~\ref{eq: forward distribution repar}
            \STATE $\mathcal{L}_\text{CT} = d(\bm{f}_{\bm{\theta}}(\bm{x}_{t}, \bm{y}_0, t),\bm{f}_{\bm{\theta}^-}(\bm{x}_{t-1}, \bm{y}_0,t-1))$
                             
            \STATE Take a gradient descent step on $\nabla_\theta \mathcal{L}_\text{CT}$
            \STATE  $k \leftarrow k +1$
        \ENDWHILE
        
        \STATE $\textbf{Stage 2: Distribution Trajectory Matching}$
        \STATE $\bm{\theta}' \leftarrow \text{stopgrad}(\bm{\theta})$
        \WHILE{not converged}
            \IF{$k \equiv 0 \pmod{1000}$}
                \STATE $\bm{f}_{\bm{\theta}'} \leftarrow \bm{f}_{\bm{\theta}}$
            \ENDIF
            \STATE sample $\bm{x}_0, \bm{y}_0 \sim (X, Y)$
            \STATE sample $t' \sim U(1, T(k))$
            \STATE compute $\bm{x}_{t'}$ using Eq.~\ref{eq: forward distribution repar}
            \STATE $\hat{\bm{x}}_0 = \bm{f}_{\bm{\theta}}(\bm{x}_{t'}, \bm{y}_0, t')$
            \STATE sample $t \sim U(T_\text{min}, T_\text{max})$
            \STATE compute $\bm{x}_{t}, \hat{\bm{x}}_{t}$ using Eq.~\ref{eq: forward distribution repar}
            \STATE $\textbf{grad}=\omega(t)(\bm{f}_{\bm{\theta}'}(\hat{\bm{x}}_t, \bm{y}_0, t) - \bm{f}_{\bm{\theta}'}(\bm{x}_t, \bm{y}_0, t)$
            
            \STATE $\mathcal{L}_{\mathrm{DTM}} = 0.5 * \text{LPIPS}(\bm{\hat{x}}_{0}, \text{stopgrad}(\bm{\hat{x}}_{0} - \textbf{grad}))$
            \STATE $\mathcal{L}_\text{total} = \lambda_\text{CT} \mathcal{L}_\text{CT} \ + \ \lambda_\text{DTM} \mathcal{L}_\text{DTM}$
            \STATE Take a gradient descent step on $\nabla_\theta \mathcal{L}_\text{total}$
            \STATE  $k \leftarrow k +1$
        \ENDWHILE
        \RETURN Converged CTMSR $f_\theta(\cdot)$.
    \end{algorithmic}
    \vspace{-0.25em}
\end{algorithm}

\begin{algorithm}[h]
    \vspace{-0.25em}
    \caption{Distribution Trajectory Matching Loss.}
    \label{alg: DTM loss}
    \begin{algorithmic}[1]
    \setstretch{1.10}
        \REQUIRE pre-trained CTMSR $\bm{f}_{\bm{\theta'}}(\cdot)$, HR image~$\bm{x}_{0}$, LR image~$\bm{y}_{0}$, timestep intervals ($T_\text{min}, T_\text{max}$), SR output $\bm{\hat{x}}_{0}$
        
        \STATE sample $t \sim U(T_\text{min}, T_\text{max})$
        \STATE compute $\bm{x}_{t}, \hat{\bm{x}}_{t}, \omega(t)$
        \STATE $\textbf{grad} = \omega(t) (\bm{f}_{\bm{\theta}'}(\hat{\bm{x}}_t, \bm{y}_0, t) - \bm{f}_{\bm{\theta}'}(\bm{x}_t, \bm{y}_0, t)$)
        \STATE $\mathcal{L}_{\mathrm{DTM}} = 0.5 * \text{LPIPS}(\bm{\hat{x}}_{0}, \text{stopgrad}(\bm{\hat{x}}_{0} - \textbf{grad}))$
        \RETURN $\mathcal{L}_{\mathrm{DTM}}$
        \vspace{-0.25em}
    \end{algorithmic}
\end{algorithm}

\vspace{-0.5em}

\subsection*{A.4. Overall Training Process}
The training process of our CTMSR can be broadly divided into two stages as mentioned in the main paper. In the first stage, we train our model exclusively with $\mathcal{L}_\text{CT}$ until convergence. Then we utilize a weighted combination of $\mathcal{L}_\text{CT}$ and $\mathcal{L}_\text{DTM}$ to further optimize our model. The total loss is formulated as:
\begin{equation}
    \mathcal{L}_\text{total} = \lambda_\text{CT} \mathcal{L}_\text{CT} \ + \ \lambda_\text{DTM} \mathcal{L}_\text{DTM}, 
\end{equation}
where we assign $\lambda_\text{CT} = 1$ and $\lambda_\text{DTM} = 1.6$. The overall training process is summarized in Algorithm~\ref{alg: overall training process}.

\section*{B. More Experimental Results}
\subsection*{B.1. Ablation Study}
To comprehensively demonstrate the effectiveness of the proposed DTM, we present additional experimental results of the ablation study on \textit{ImageNet-Test}, RealSet65 and RealSR datasets. The results demonstrate the effectiveness of DTM across synthetic and real-world datasets. The detailed results are shown in Table~\ref{tab: ablation imagenet},~\ref{tab: ablation realset},~\ref{tab: ablation realsr}. 

\subsection*{B.2. Compared with SinSR}
\begin{figure}[t]
    \centering
    \includegraphics[width=0.45\textwidth,keepaspectratio]{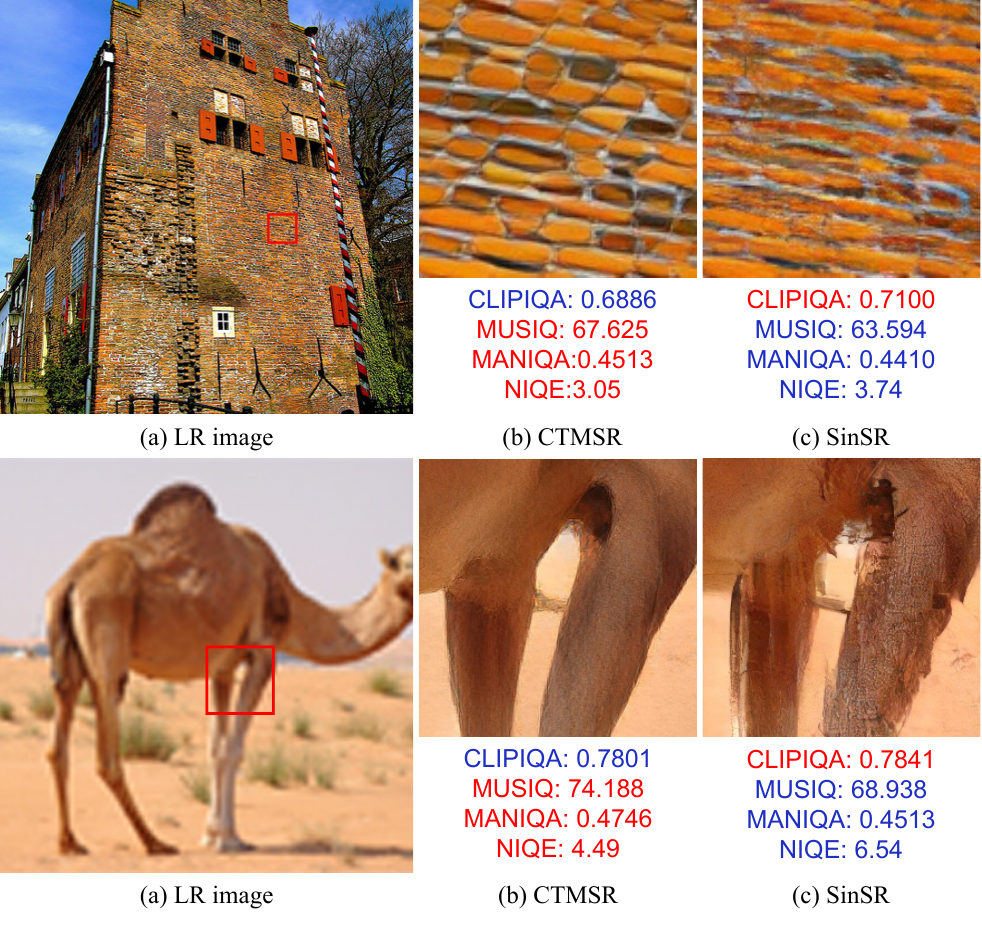}
    \captionsetup{width=0.45\textwidth}
    \vspace{-1em}
    \caption{An illustration of CLIPIQA's tendency to favor images with noise or artifacts and its inability to effectively distinguish between fine image details and noise or artifacts. Here are two visual examples of CTMSR and SinSR.}
    \label{pic: sup_compare_sinsr}
    \vspace{-1em}
\end{figure}

The test results on RealSet65 and RealSR (shown in Table~\ref{tab: real_test}) demonstrate that our method outperforms SinSR~\cite{wang2024sinsr} across all metrics except for CLIPIQA. Upon detailed observation, we discover that the CLIPIQA tends to favor images with noise or artifacts and sometimes fails to distinguish between fine image details and noise or artifacts. Therefore, CLIPIQA occasionally produces higher scores for images of lower quality due to the presence of noise or artifacts. The visual examples are shown in Figure~\ref{pic: sup_compare_sinsr}.

\subsection*{B.3. Compared with Stable Diffusion-Based Methods}
Though Stable Diffusion-based methods achieve impressive results, they rely on the powerful generative capabilities of Stable Diffusion (SD). This results in these methods being constrained by fixed backbones (Stable Diffusion), which limits their scalability to smaller models and consequently restricts their applicability in practical scenarios. 
In addition, these methods require extremely large models and incur significant inference costs, placing them in a different track from our approach. To compare with SD-based methods, we apply our approach to the latent space provided by VQ-VAE to further enhance the performance of our model. As shown in Table~\ref{tab: compare_sd}, our refined method attains performance on par with SD-based methods with much fewer model parameters and inference time. To be more specific, 
(1) OSEDiff demands \textbf{1.7} times the inference time and \textbf{8} times the number of model parameters; 
(2) AddSR demands \textbf{3.7} times the inference time and \textbf{10} times the number of model parameters.

\subsection*{B.3. Visual Comparison}
\begin{figure*}[t]
    \centering
    \includegraphics[width=0.90\textwidth,keepaspectratio]{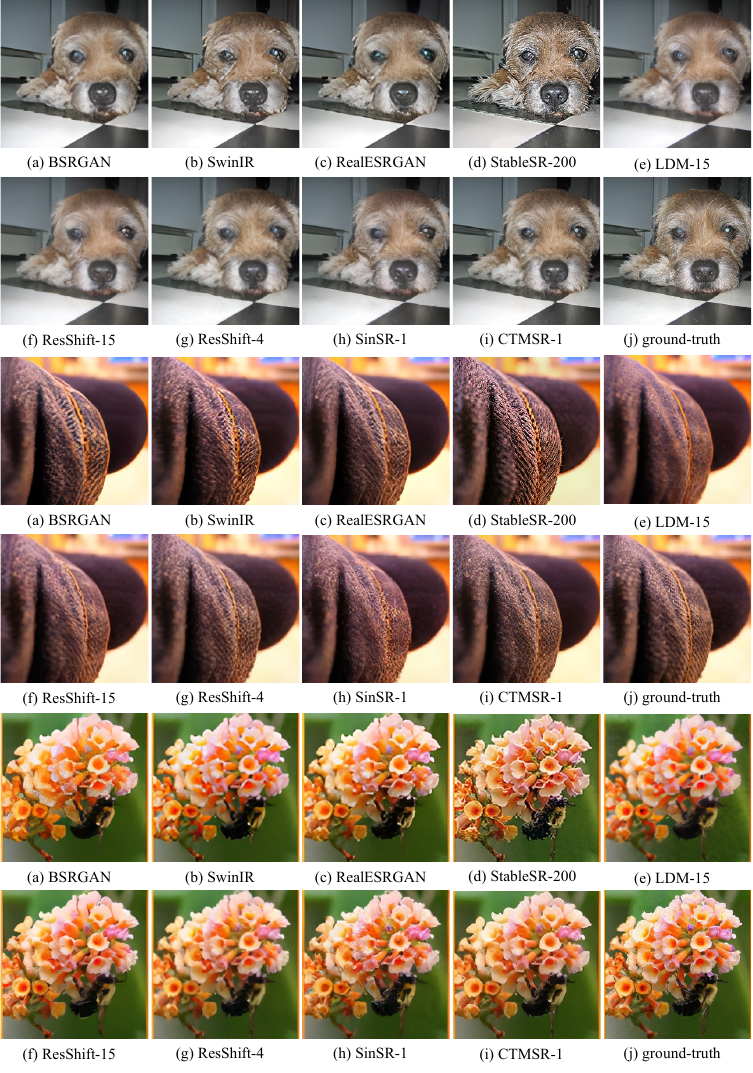}
    \captionsetup{width=0.90\textwidth}
    \vspace{-1em}
    \caption{Visual comparison of different methods on \textit{ImageNet-Test}. Please zoom in for more details. }
    \label{pic: visual_sup_imagenet_1}
    \vspace{-1em}
\end{figure*}

\begin{figure*}[t]
    \centering
    \includegraphics[width=0.90\textwidth,keepaspectratio]{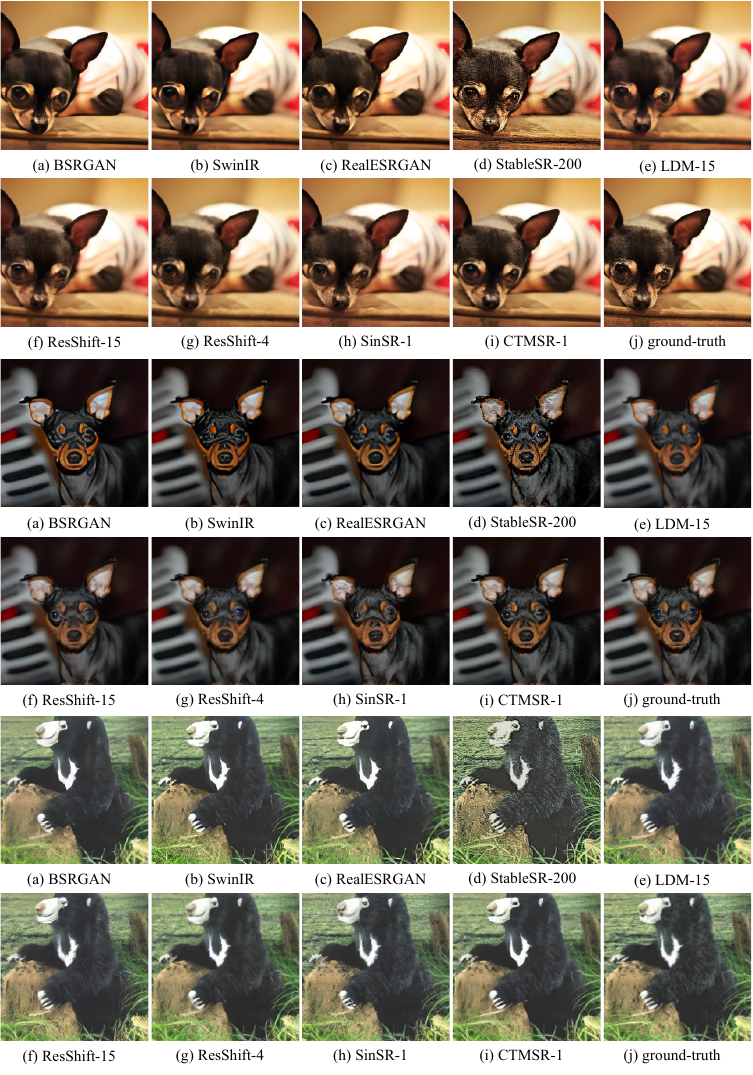}
    \captionsetup{width=0.90\textwidth}
    \vspace{-1em}
    \caption{Visual comparison of different methods on \textit{ImageNet-Test}. Please zoom in for more details. }
    \label{pic: visual_sup_imagenet_2}
    \vspace{-1em}
\end{figure*}

\begin{figure*}[t]
    \centering
    \includegraphics[width=0.89\textwidth,keepaspectratio]{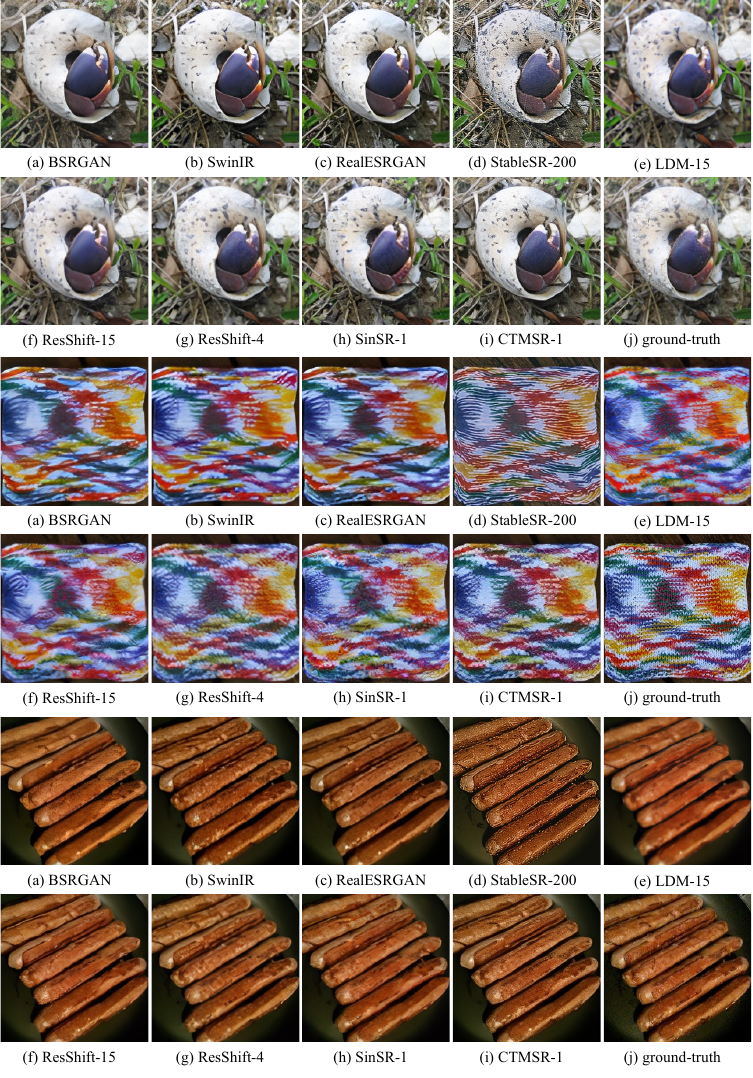}
    \captionsetup{width=0.90\textwidth}
    \vspace{-1em}
    \caption{Visual comparison of different methods on \textit{ImageNet-Test}. Please zoom in for more details. }
    \label{pic: visual_sup_imagenet_3}
    \vspace{-1em}
\end{figure*}

\begin{figure*}[t]
    \centering
    \includegraphics[width=0.90\textwidth,keepaspectratio]{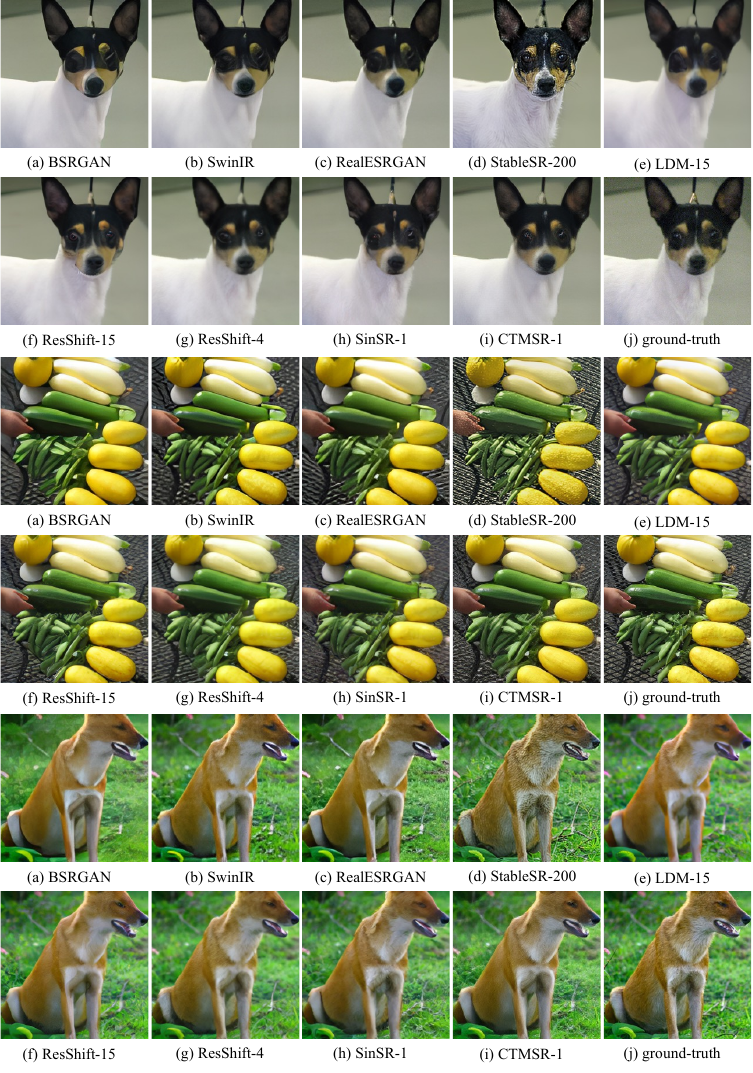}
    \captionsetup{width=0.90\textwidth}
    \vspace{-1em}
    \caption{Visual comparison of different methods on \textit{ImageNet-Test}. Please zoom in for more details. }
    \label{pic: visual_sup_imagenet_4}
    \vspace{-1em}
\end{figure*}

\begin{figure*}[t]
    \centering
    \includegraphics[width=0.95\textwidth,keepaspectratio]{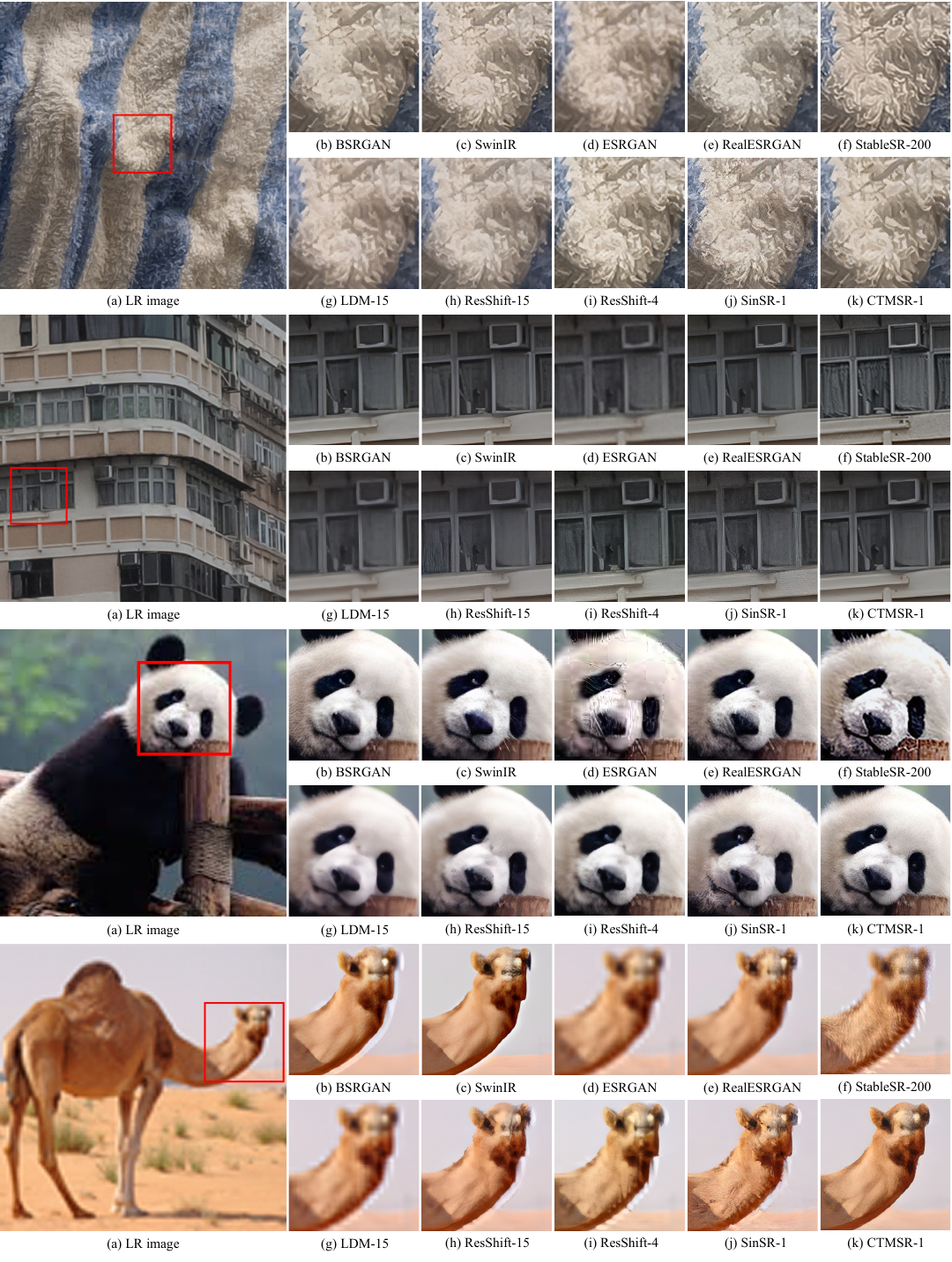}
    \captionsetup{width=0.95\textwidth}
    \vspace{-1em}
    \caption{Visual comparison of different methods on real-world datasets. Please zoom in for more details. }
    \label{pic: visual_sup_real_1}
    \vspace{-1em}
\end{figure*}

\begin{figure*}[t]
    \centering
    \includegraphics[width=0.95\textwidth,keepaspectratio]{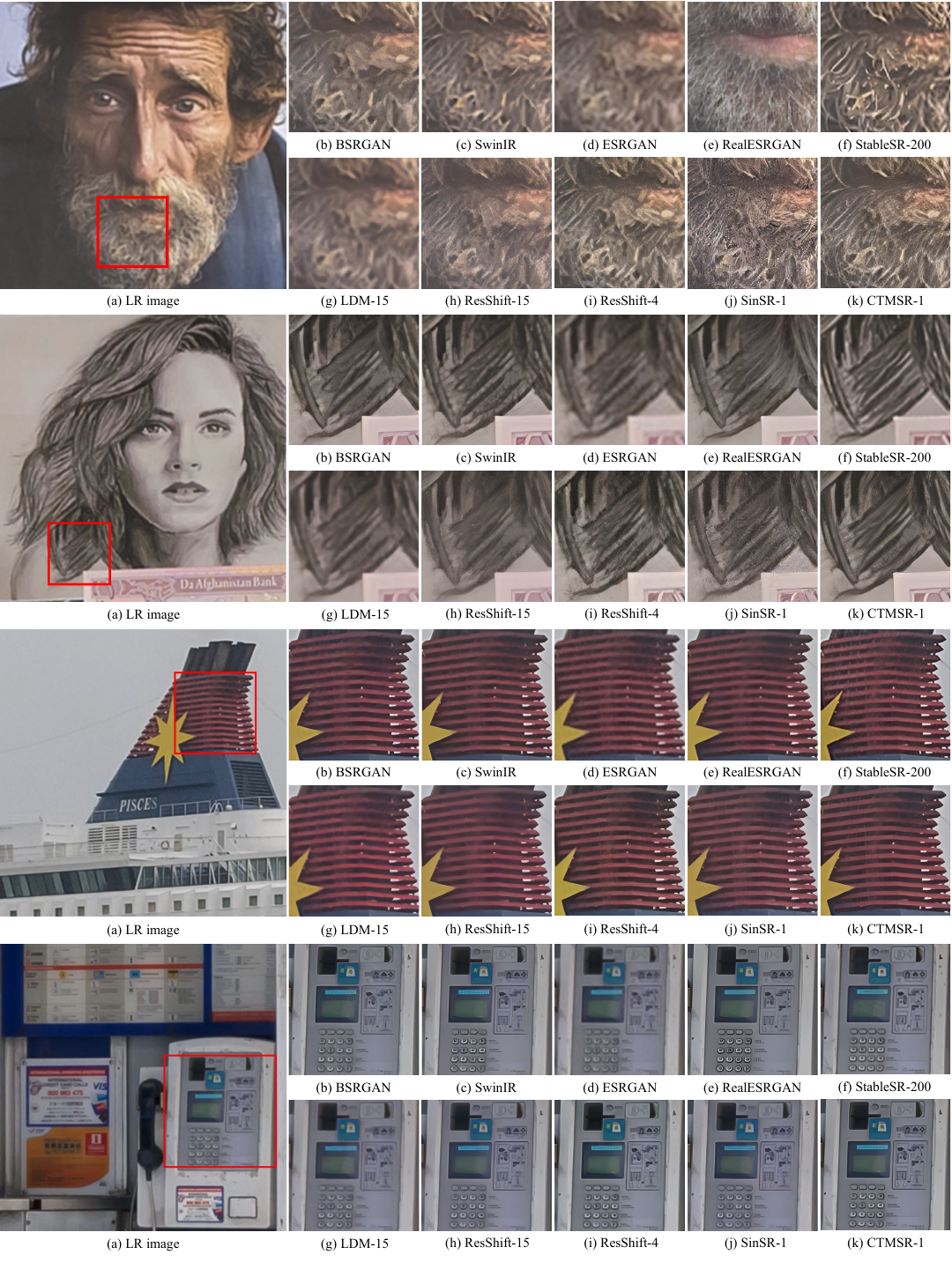}
    \captionsetup{width=0.95\textwidth}
    \vspace{-1em}
    \caption{Visual comparison of different methods on real-world datasets. Please zoom in for more details. }
    \label{pic: visual_sup_real_2}
    \vspace{-1em}
\end{figure*}

\begin{figure*}[t]
    \centering
    \includegraphics[width=0.95\textwidth,keepaspectratio]{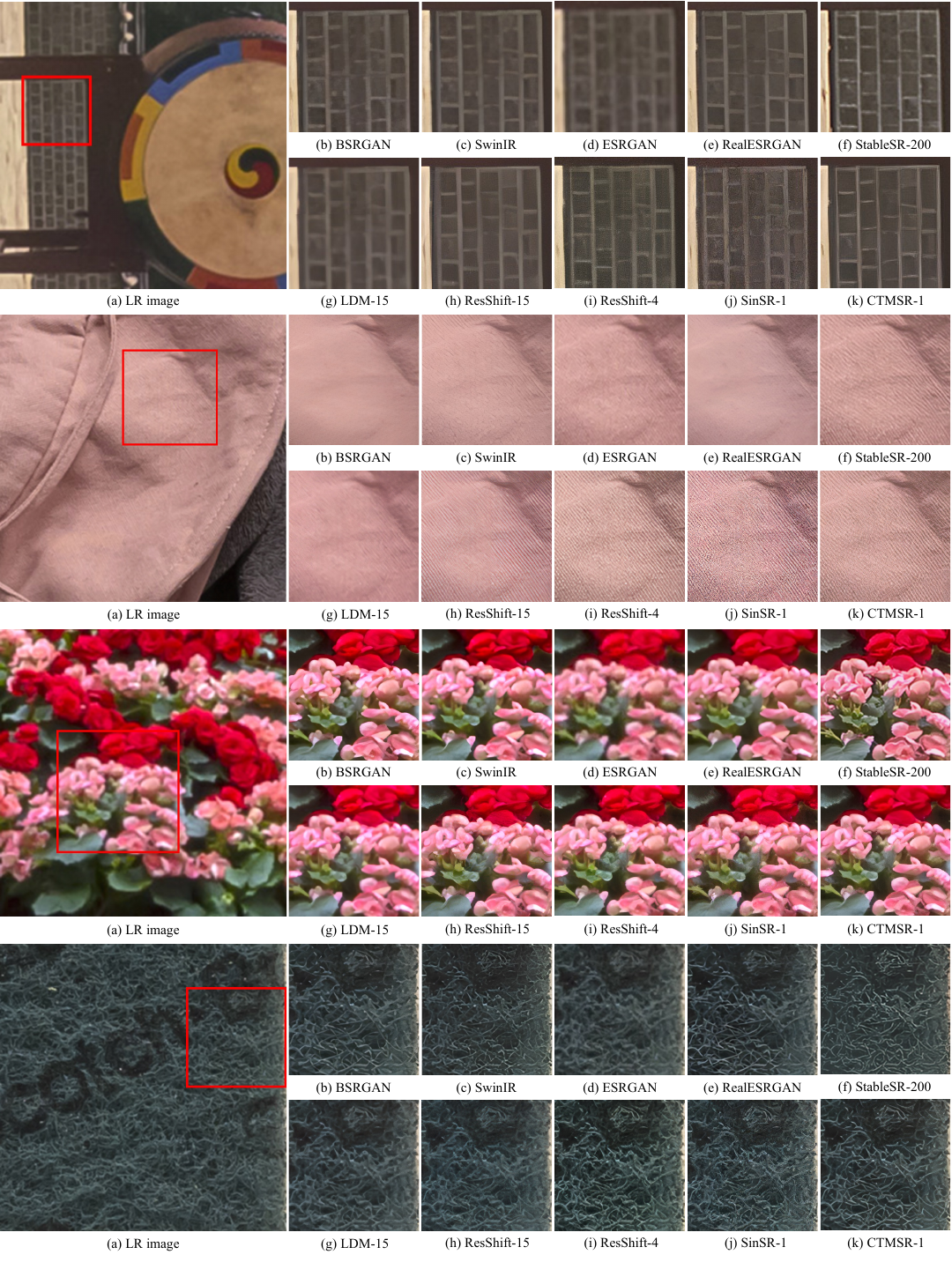}
    \captionsetup{width=0.95\textwidth}
    \vspace{-1em}
    \caption{Visual comparison of different methods on real-world datasets. Please zoom in for more details. }
    \label{pic: visual_sup_real_3}
    \vspace{-1em}
\end{figure*}

We provide more visual examples of CTMSR compared with recent state-of-the-art methods on \textit{ImageNet-Test} and real-world datasets. The visual examples are shown in Figure~\ref{pic: visual_sup_imagenet_1}, ~\ref{pic: visual_sup_imagenet_2}, ~\ref{pic: visual_sup_imagenet_3}, ~\ref{pic: visual_sup_imagenet_4} ~\ref{pic: visual_sup_real_1}, ~\ref{pic: visual_sup_real_2}, ~\ref{pic: visual_sup_real_3}.

\begin{table}
\footnotesize
\setlength{\tabcolsep}{5pt}



\begin{center}
\begin{tabular}{@{}C{2.5cm}@{}|@{}C{1.2cm}@{}@{}C{1.2cm}@{}@{}C{1.5cm}@{}@{}C{1.5cm}@{}}
\Xhline{0.8pt}
\multicolumn{1}{c|}{Methods} & PSNR$\uparrow$ & LPIPS$\downarrow$ & CLIPIQA$\uparrow$ & MUSIQ$\uparrow$ \\  
\Xhline{0.4pt}

CTMSR (w/o DTM) & 24.71 &  0.2004 & 0.6092 & 56.650\\
CTMSR (w/ SDS) & 23.17 & 0.2545 & 0.6292 & 58.188 \\
CTMSR (w/ DTM) & \textbf{24.73} & \textbf{0.1969} & \textbf{0.6913} & \textbf{60.142} \\

\Xhline{0.8pt}
\end{tabular}
\vspace{-2mm}
\caption{Experimental results of ablation study on \textit{ImageNet-Test}.}
\label{tab: ablation imagenet}
\vspace{-1em}
\end{center}
\vspace{-0.5em}
\end{table}

\begin{table}
\footnotesize
\setlength{\tabcolsep}{5pt}



\begin{center}
\begin{tabular}{@{}C{2.5cm}@{}|@{}C{1.5cm}@{}@{}C{1.5cm}@{}@{}C{1.5cm}@{}@{}C{1.2cm}@{}}
\Xhline{0.8pt}
\multicolumn{1}{c|}{Methods} & CLIPIQA$\uparrow$ & MUSIQ$\uparrow$ & MANIQA$\uparrow$ & NIQE$\downarrow$ \\  
\Xhline{0.4pt}

CTMSR (w/o DTM) & 0.6009 &  64.274 & 0.3658 & \textbf{4.37}\\
CTMSR (w/ SDS) & 0.6446 & 62.217 & 0.3606 & 4.77 \\
CTMSR (w/ DTM) & \textbf{0.6893} & \textbf{67.173} & \textbf{0.4360} & 4.51 \\

\Xhline{0.8pt}
\end{tabular}
\vspace{-2mm}
\caption{Experimental results of ablation study on RealSet65.}
\label{tab: ablation realset}
\vspace{-1em}
\end{center}
\vspace{-0.5 em}
\end{table}

\begin{table}
\footnotesize
\setlength{\tabcolsep}{5pt}



\begin{center}
\begin{tabular}{@{}C{2.5cm}@{}|@{}C{1.5cm}@{}@{}C{1.2cm}@{}@{}C{1.5cm}@{}@{}C{1.2cm}@{}}
\Xhline{0.8pt}
\multicolumn{1}{c|}{Methods} & CLIPIQA$\uparrow$ & MUSIQ$\uparrow$ & MANIQA$\uparrow$ & NIQE$\downarrow$ \\  
\Xhline{0.4pt}

CTMSR (w/o DTM) & 0.5542 & 62.351 & 0.3512 & \textbf{4.33}\\
CTMSR (w/ SDS) & 0.6101 & 60.919 & 0.3479 & 5.11 \\

CTMSR (w/ DTM) & \textbf{0.6449} & \textbf{64.796} & \textbf{0.4157} & 4.65 \\

\Xhline{0.8pt}
\end{tabular}
\vspace{-2mm}
\caption{Experimental results of ablation study on RealSR.}
\label{tab: ablation realsr}
\vspace{-1em}
\end{center}
\vspace{-1em}
\end{table}

\begin{table}[t]
\footnotesize
\setlength{\tabcolsep}{5pt}



\begin{center}
\begin{tabular}{@{}C{1.2cm}@{}|@{}C{1.5cm}@{}@{}C{1.5cm}@{}|@{}C{1.3cm}@{}@{}C{1.2cm}@{}@{}C{1.3cm}@{}}
\Xhline{0.8pt}
{Methods}& Runtime (s) & {Params (M)}& CLIPIQA$\uparrow$ & MUSIQ$\uparrow$ & MANIQA$\uparrow$ \\  
\Xhline{0.4pt}
OSEDiff & 0.3100 & 1775 & 0.6693 & \textbf{69.10} & 0.4717  \\
AddSR & 0.6857 & 2280 & 0.5410 & 63.01 & 0.4113 \\
CTMSR & \textbf{0.1847} & \textbf{225} & \textbf{0.7420} & 64.81 & \textbf{0.4810} \\

\Xhline{0.8pt}
\end{tabular}
\vspace{-2mm}
\caption{Quantitative comparisons with SD-based methods on RealSR. The runtime is tested on $128\times 128$ input images.}
\label{tab: compare_sd}
\vspace{-1em}
\end{center}
\vspace{-1em}
\end{table}

